\documentclass{article}

\usepackage[main, final]{neurips_2026}
\usepackage[utf8]{inputenc} % allow utf-8 input
\usepackage[T1]{fontenc}    % use 8-bit T1 fonts
\usepackage{hyperref}       % hyperlinks
\usepackage{url}            % simple URL typesetting
\usepackage{booktabs}       % professional-quality tables
\usepackage{amsfonts}       % blackboard math symbols
\usepackage{nicefrac}       % compact symbols for 1/2, etc.
\usepackage{microtype}      % microtypography
\usepackage{xcolor}         % colors
\usepackage{graphicx}
\usepackage{subcaption}
\usepackage{array}
\usepackage{amsmath}
\usepackage{multirow} 

\title{AmbientFM: A Foundation Model for Ambient Sensing}

\author{%
  \textbf{Guozhen Zhu}$^{1}$\quad
  \textbf{Yuqian Hu}$^{1}$\quad
  \textbf{Sakila S Jayaweera}$^{1}$\quad
  \textbf{Wei-Hsiang Wang}$^{1}$\quad
  \textbf{Weihang Gao}$^{1}$\\
  \textbf{Jiaxuan Zhang}$^{1}$\quad
  \textbf{Beibei Wang}$^{1}$\quad
  \textbf{Chenshu Wu}$^{2}$\quad
  \textbf{K. J. Ray Liu}$^{1}$\\[0.45em]
  $^{1}$ADT Research\qquad
  $^{2}$The University of Hong Kong
}
\begin{document}

\maketitle

\begin{abstract}
Ambient intelligence seeks to continuously understand human presence, activity, and physiology in physical spaces, enabling smart environments, health monitoring, and human-computer interaction. WiFi infrastructure offers a ubiquitous, always-on, and privacy-preserving sensing substrate across billions of IoT devices. Yet ambient sensing with WiFi remains largely fragmented, with most systems relying on task-specific models, labeled data collection, and customized training pipelines.

We present AmbientFM, a foundation model for ambient sensing through ubiquitous WiFi signals. AmbientFM is pre-trained on 9.2 million unlabeled Channel State Information (CSI) samples collected over 439 days from 20 commercial device types deployed in real homes, the largest real-world WiFi CSI pre-training corpus reported to date. It learns transferable wireless representations through contrastive learning, masked reconstruction, and physics-informed objectives tailored to wireless signals. With lightweight adaptation, a single backbone transfers to 9 downstream tasks, achieving above 0.90 AUROC on all classification benchmarks and supporting dense spatial reconstruction. AmbientFM reduces the need for labeled data and task-specific design, enabling scalable ambient intelligence on existing wireless infrastructure.
\href{https://ai-sensing.github.io/ambient-fm/index.html}{\textcolor{blue}{\textit{Project Page.}}}
\end{abstract}

%%%%%%%%%%%%%%%%%%%%%%%%%%%%%%%%%%%%%%%%%%%%%%%%%%%%%%%%%%%%%%%%%%%%%%%%%%%%%%%
% INTRODUCTION
%%%%%%%%%%%%%%%%%%%%%%%%%%%%%%%%%%%%%%%%%%%%%%%%%%%%%%%%%%%%%%%%%%%%%%%%%%%%%%%
\section{Introduction}
\label{sec:intro}

% Ambient intelligence that continuously senses and responds to its occupants underpins applications from smart home automation to elderly care and occupational health monitoring. Realizing this vision requires perception that is always-on, unobtrusive, and deployable at scale. 
% Ambient WiFi signals offer a compelling substrate for ambient perception. They propagate through walls, function in any lighting condition, require no contact or worn devices, and preserve visual privacy. The billions of access points, smart speakers, and IoT devices already deployed continuously transmit signals that interact with the environment and its occupants, and their channel state information (CSI) captures motion patterns as well as subtle physiological signals such as respiration. This ubiquitous wireless substrate could enable ambient perception at unprecedented scale without additional infrastructure.

Ambient intelligence that continuously senses and responds to its occupants underpins applications from smart home automation to elderly care and occupational health monitoring. Realizing this vision requires perception that is always-on, unobtrusive, and deployable at scale. Ambient WiFi signals offer a compelling substrate: they propagate through walls, function in any lighting condition, require no worn devices, and preserve visual privacy \cite{wificandomore}. The billions of access points, smart speakers, and IoT devices already deployed continuously transmit signals whose channel state information (CSI) captures human motion patterns and subtle physiological signals such as respiration, enabling ambient perception at unprecedented scale without additional hardware.

% Despite this potential, ambient sensing research has predominantly relied on task-specific models trained from scratch for individual applications. Each new task or sensing scenario requires designing custom architectures, collecting labeled data, and training specialized models. This fragmentation limits practical deployment: substantial labeled data is needed for every task and setting, and knowledge cannot be shared across the diverse sensing capabilities that WiFi infrastructure enables.

Despite this potential, ambient sensing research has predominantly relied on task-specific models trained from scratch. Each new task requires custom architectures, large-scale labeled data collection, and specialized training, and knowledge cannot be shared across the diverse sensing capabilities that WiFi enables \cite{WiFisurvey_FM}. Foundation models have shown that large-scale self-supervised pre-training produces general-purpose representations that transfer across tasks and reduce labeling requirements in vision~\cite{MAE,CLIP}, speech~\cite{wav2vec}, and inertial sensing~\cite{IMU-FM}. WiFi is a natural candidate for this paradigm: commodity hardware already generates massive unlabeled CSI streams, and a single backbone could serve applications from room-level occupancy to fine-grained gesture recognition.

% Foundation models have demonstrated that large-scale self-supervised pre-training on diverse data produces general-purpose representations that transfer effectively across tasks and domains. Drawing from the success of foundation models in natural language processing and computer vision, we explore whether this paradigm extends to ambient perception. WiFi presents a compelling opportunity for this approach: commodity hardware already generates massive streams of unlabeled CSI data; a single pre-trained backbone can serve applications from coarse spatial reasoning to fine-grained physiological monitoring; and WiFi signals penetrate walls, enabling ambient sensing in conditions where other modalities struggle. 

% However, extending foundation models to ambient sensing via WiFi presents unique technical challenges. CSI exhibits non-local frequency structure arising from multipath propagation, which is distinct from the spatial locality observed in images or the spectral smoothness characteristic of audio signals. Subcarriers have heterogeneous signal quality depending on hardware, interference, and propagation conditions, requiring adaptive processing across the frequency dimension. Human activities induce characteristic patterns that benefit from translation-invariant temporal modeling. Additionally, the modality lacks established large-scale pre-training datasets, and standard self-supervised objectives designed for vision or language may not capture the physics of wireless signal propagation.

However, WiFi is a communication system repurposed for sensing, and this origin introduces challenges that are qualitatively more severe than those in purpose-built sensing modalities. WiFi devices dynamically change their bandwidth, frequency band, and antenna configuration at runtime, producing structurally incompatible observations rather than simple calibration differences. The MAC-layer protocol introduces non-deterministic packet timing and abrupt amplitude shifts, while omnidirectional antennas entangle human motion with environment-specific multipath structure that typically dominates the observed signal. These properties render standard self-supervised objectives insufficient, necessitating methods tailored to wireless signals.

In this paper, we present AmbientFM\footnote{Project webpage: \url{https://ai-sensing.github.io/ambient-fm/index.html}}, a foundation model for WiFi-based ambient intelligence, pre-trained at scale on in-the-wild unlabeled CSI data and designed for efficient adaptation across tasks. It is pretrained on 9.2M unlabeled CSI samples collected over 439 days from 20 device types across 11 environments, to our knowledge the largest real-world corpus used for WiFi CSI pre-training. Its self-supervised framework combines contrastive learning, masked reconstruction, and a physics-informed autocorrelation objective. The architecture incorporates adaptive frequency aggregation to handle heterogeneous subcarrier quality and relative temporal encoding for translation-invariant activity patterns, and is evaluated across 9 downstream tasks via parameter-efficient adaptation.

% Our key contributions include: 1) \textbf{Developing a foundation model for ambient intelligence}: We outline the design and development of the first large-scale foundation model for ambient sensing via WiFi.  2) \textbf{Domain-specific pre-training with large-scale real-world dataset}: We show that with a large-scale unlabeled WiFi dataset, a self-supervised learning framework tailored to the characteristics of wireless signals can extract representations for practical ambient perception. 3) \textbf{Strong performance of single backbone on diverse tasks with superior data efficiency}: We demonstrate AmbientFM's generalizability and real-world utility across 9 diverse WiFi sensing tasks from public benchmarks, establishing that the foundation model paradigm extends successfully to WiFi-based ambient perception.

Our key contributions include: 1) \textbf{Developing a foundation model for ambient intelligence}: We outline the design and development of a large-scale foundation model for ambient sensing based on WiFi, providing a single backbone for diverse ambient-sensing tasks. 2) \textbf{Domain-specific pre-training with large-scale real-world dataset}: We show that with a large-scale unlabeled CSI dataset, a self-supervised learning framework tailored to the characteristics of wireless signals can extract representations for practical ambient perception. 3) \textbf{Strong performance of single backbone on diverse tasks with superior data efficiency}: We demonstrate AmbientFM's generalizability and real-world utility across 9 diverse WiFi sensing tasks from public benchmarks spanning physiological sensing, activity understanding, spatial reasoning, biometrics, fine-grained motion, and spatial reconstruction. AmbientFM achieves strong performance with lightweight adaptation and reduces the need for task-specific data and design, offering a practical path toward scalable, privacy-preserving ambient intelligence on existing wireless infrastructure.

\begin{figure*}
    \centering
    \includegraphics[width=1\linewidth]{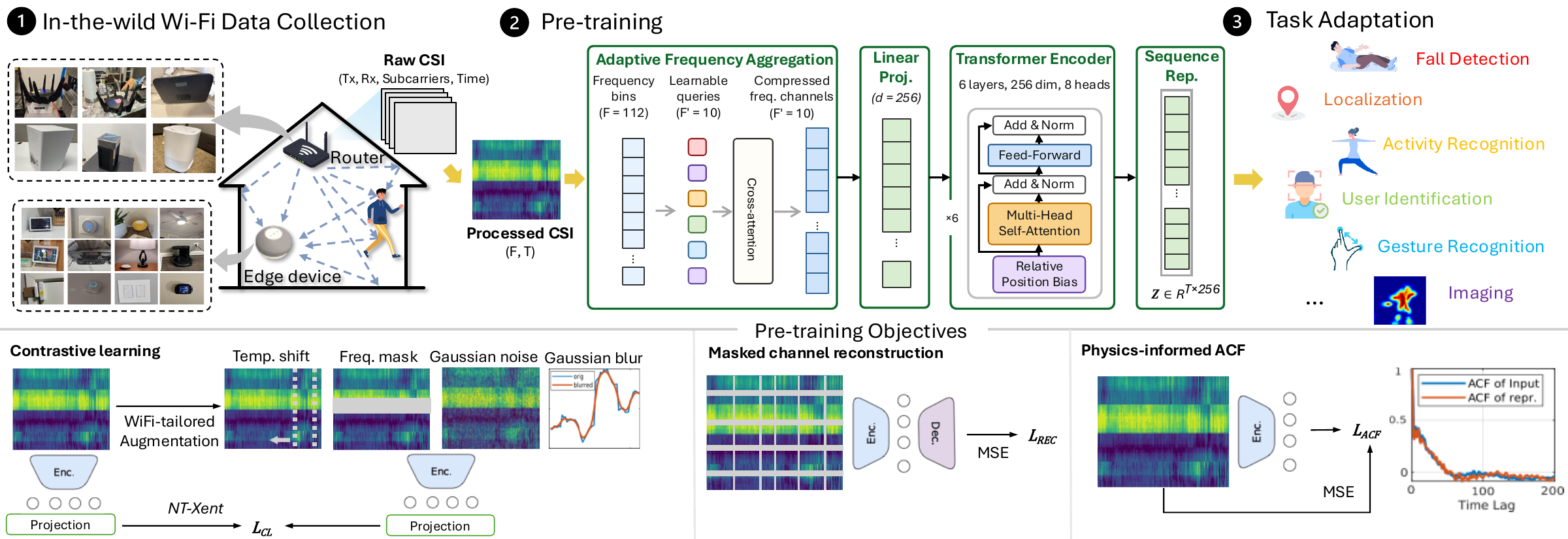}
    \caption{\textbf{AmbientFM overview.} AmbientFM pre-trains a shared sensing backbone on in-the-wild CSI with complementary self-supervised objectives, then adapts it to diverse downstream tasks with lightweight tuning.}
    \label{fig:overview}
\end{figure*}

%%%%%%%%%%%%%%%%%%%%%%%%%%%%%%%%%%%%%%%%%%%%%%%%%%%%%%%%%%%%%%%%%%%%%%%%%%%%%%%
% RELATED WORK
%%%%%%%%%%%%%%%%%%%%%%%%%%%%%%%%%%%%%%%%%%%%%%%%%%%%%%%%%%%%%%%%%%%%%%%%%%%%%%%
\section{Related Work}
\label{sec:related}

\textbf{Foundation models for perception.}
Foundation models have transformed representation learning by enabling transfer from large-scale unlabeled data. In vision, masked autoencoders and contrastive learning produce transferable representations under limited supervision~\cite{MAE, CLIP, SimCLR, MoCo}. This paradigm extends to other sensing modalities: wav2vec 2.0 improves sample efficiency in speech recognition~\cite{wav2vec}, and pre-trained models for inertial sensing demonstrate cross-user and cross-device transfer~\cite{IMU-FM}. A consistent finding is that foundation models reduce labeling requirements while enabling knowledge sharing across tasks. For WiFi-based ambient sensing, this paradigm has only recently begun to be explored, as discussed below.

\textbf{WiFi sensing capabilities and limitations.}
WiFi CSI captures how radio signals propagate through indoor environments, enabling diverse applications including activity recognition~\cite{activity-1, activity-3}, fall detection~\cite{fall-2,fall-1}, gesture detection~\cite{gesture-1,gesture-2}, respiration monitoring~\cite{respiration-1,respiration-2}, and indoor localization~\cite{localization-1,localization-3}. WiFi signals propagate through walls, require no wearable devices, and leverage billions of already deployed access points. However, current research relies on task-specific models trained from scratch~\cite{fall,localization-2,localization-3}, with each new task requiring custom architectures and labeled data. Prior efforts in domain adaptation~\cite{CrossSense}, adversarial learning~\cite{EI}, and physics-driven feature engineering~\cite{Widar3.0} improve transferability for individual tasks but do not provide general-purpose representations. Whether a single pre-trained model can support multiple WiFi sensing tasks remains largely unexplored.

\textbf{Self-supervised and foundation-model approaches for wireless signals.}
SSL has improved sample efficiency in WiFi sensing~\cite{SSL}, but existing methods pre-train on small, task-specific datasets and evaluate one or two tasks, mostly activity recognition~\cite{SSL_HAR_TOSN,CIG_MAE,CAPC}. Wireless foundation models such as WiFo-CF~\cite{WiFoCF} and CSI-MAE~\cite{CSI_MAE} pre-train on simulated channels for communication-layer problems, including CSI feedback and channel modeling, whose objectives and outputs differ from human-centric sensing. Concurrent work has also explored predictive pre-training for WiFi sensing~\cite{CSI_JEPA,WiFi_JEPA}. AmbientFM differs from these efforts in pre-training a task-agnostic backbone on large-scale real-world CSI from heterogeneous commodity devices and in evaluating cross-task transfer across nine ambient-sensing tasks (Table~\ref{tab:related}).

\begin{table}[t]
\centering
\small
\caption{Comparison with related self-supervised and foundation-model work on wireless signals.}
\label{tab:related}
\resizebox{\linewidth}{!}{%
\begin{tabular}{@{}llcccc@{}}
\toprule
\textbf{Method} & \textbf{Application domain} & \textbf{Foundation model} & \textbf{Pre-training scale} & \textbf{Real/Sim.} & \textbf{\#Tasks} \\
\midrule
SSL-HAR~\cite{SSL_HAR_TOSN} & Sensing & No & $\le$45K / setting & Real & 1 \\
CIG-MAE~\cite{CIG_MAE} & Sensing & No & 2.2K--3.8K / dataset & Real & 1 \\
CAPC~\cite{CAPC} & Sensing & No & 5.5K & Real & 2 \\
CSI-MAE~\cite{CSI_MAE} & Wireless channel modeling & Yes & 1.45M CSI & Sim. & 3 \\
WiFo-CF~\cite{WiFoCF} & Wireless communication & Yes & 160K CSI & Sim. & 2 \\
\midrule
AmbientFM & Ambient sensing with WiFi & Yes & 9.2M / 439 days & Real & 9 \\
\bottomrule
\end{tabular}}
\vspace{-0.5em}
\end{table}

% The proposed \textbf{AmbientFM (WiFi Foundation Model)} is developed to learn general-purpose, robust representations from large-scale WiFi sensing data. It supports a wide range of downstream tasks through minimal supervision. This section describes the complete methodology, which includes data preprocessing, self-supervised pretraining, model architecture design, and multi-task adaptation.

% The framework of the proposed AmbientFM is illustrated in Figure \ref{fig:overview}, 

\section{Data}
\label{sec:data}

\subsection{Pre-training Dataset}
\label{sec:pretraining_dataset}

We collect a large-scale in-the-wild WiFi CSI corpus for self-supervised pre-training, comprising 439+ days of recording, \(\sim\)20~TB of data, and 9.2 million CSI samples across 20 commercial IoT device types, 33 physical devices, 11 real-world environments, and 26 participants. To our knowledge, this is the largest and most diverse WiFi sensing dataset reported to date for foundation model pre-training.

The corpus spans substantial hardware and deployment diversity, including 8 chipset families, \(1\times1\) to \(2\times2\) MIMO, 20/40/80~MHz bandwidths, and both 2.4 and 5~GHz bands, with more than 20~dB variation in received power across devices (Appendix~\ref{app:device_summary}). The environments include studios, apartments, houses, and townhouses with diverse layouts, scales, and interference conditions (Appendix~\ref{app:env}). CSI is recorded continuously during natural daily activities with 6--9 concurrent device links per environment, capturing realistic motion, occlusion, multipath, and interference beyond lab settings. Appendix~\ref{app:dataset}, Table~\ref{tab:corpus_compare} compares the corpus with representative public WiFi sensing datasets.
 
\subsection{WiFi CSI: Structure and Challenges}
\label{sec:csi_properties}

CSI describes how wireless signals propagate between a transmitter and a receiver by measuring the amplitude and phase responses of frequency subcarriers. When people or objects move in the environment, they perturb these channel responses, allowing CSI to support sensing applications such as activity recognition, localization, and physiological monitoring~\cite{survey1}.

CSI is a rich sensing signal, but its structure differs from  measurement produced by dedicated sensors. Because WiFi is designed primarily for communication, devices may adapt their bandwidth, frequency band, antenna configuration, and transmission power at runtime, sometimes even within a single session due to communication demands, making the signal structure like subcarriers inherently unpredictable for sensing model~\cite{zhu2025experience}. The MAC-layer protocol also introduces irregular packet timing and transmission behavior optimized for communication. Automatic Gain Control can further introduce abrupt amplitude changes. In addition, commodity WiFi uses omnidirectional antennas, so motion-induced changes are observed together with environment-specific multipath, whose static components can be 10--100$\times$ stronger than the motion perturbations~\cite{robust_proximity}. These properties make CSI highly informative, but they also require representations that can handle hardware diversity, protocol effects, and environment-dependent propagation.

These signal properties directly guide AmbientFM's design. First, multipath propagation creates non-local and configuration-dependent correlations across distant subcarriers, which vary with bandwidth and band selection~\cite{subcarrierselect}. To model this structure, AmbientFM uses adaptive frequency aggregation to compress heterogeneous subcarrier inputs into a unified representation. Second, subcarrier quality varies across hardware, packet timing, and frequency-selective fading~\cite{mrc1,subcarrierselect}. AmbientFM addresses this with learned subcarrier weighting and masked reconstruction, which exposes the model to packet loss and frequency-selective corruption during pre-training. Third, human activities produce characteristic temporal patterns and these patterns are embedded in measurements dominated by static multipath, AmbientFM uses relative temporal encoding and a physics-informed Autocorrelation Function (ACF) objective to emphasize motion-induced temporal structure while reducing sensitivity to static environmental effects~\cite{WiDetect,WiSpeed,frequency1,frequency2}.
 
\subsection{Data Preprocessing}
\label{sec:preprocessing}
 
Raw CSI is parsed into a complex-valued tensor $\mathbf{H} \in \mathbb{C}^{N_{\text{tx}} \times N_{\text{rx}} \times N_{\text{sub}} \times T}$, where $N_{\text{tx}}, N_{\text{rx}}$ denote antenna counts, $N_{\text{sub}}$ is the number of subcarriers (varying with channel bandwidth), and $T$ is temporal length. We flatten spatial--frequency dimensions to $\mathbf{X} \in \mathbb{C}^{F \times T}$ with $F = N_{\text{tx}} N_{\text{rx}} N_{\text{sub}}$. CSI streams are collected at a target rate of 100~Hz; we estimate effective rates from inter-packet timestamps and discard streams below two-thirds of the target rate.
 
We extract CSI amplitude $\mathbf{A} = |\mathbf{X}|$, which is robust to the phase noise common in commodity hardware~\cite{phase1}; the eight chipset families in our corpus also do not share a common phase reference. We segment continuous streams into non-overlapping fixed-length windows, producing segments $\mathbf{S}_i \in \mathbb{R}^{T \times F}$ with $T=500$, i.e., 5\,s at the 100~Hz target rate. All segments are zero-padded to a unified dimension $F_{\max}$ defined by the maximum feature size across devices, normalized to $[0,1]$ via per-segment min--max normalization, and stored in compressed HDF5 format. No task-driven segmentation or filtering is applied.

%%%%%%%%%%%%%%%%%%%%%%%%%%%%%%%%%%%%%%%%%%%%%%%%%%%%%%%%%%%%%%%%%%%%%%%%%%%%%%%
% SECTION 4: FOUNDATION MODEL DEVELOPMENT
%%%%%%%%%%%%%%%%%%%%%%%%%%%%%%%%%%%%%%%%%%%%%%%%%%%%%%%%%%%%%%%%%%%%%%%%%%%%%%%
\section{Foundation Model Development}
\label{sec:foundation_model}
 
We present a self-supervised framework for learning transferable WiFi CSI representations. Our design choices are guided by the structural properties of CSI identified in Section~\ref{sec:csi_properties}, with the goal of learning representations that generalize across a wide variety of sensing tasks.
 
\subsection{Architecture}
\label{sec:architecture}

The encoder maps raw CSI segments into compact representations using two components: adaptive frequency aggregation and relative temporal encoding.

To address variation in subcarrier quality across devices and environments, we use cross-attention to compress $F$ input subcarriers into $F' = 10$ latent frequency channels:
\begin{equation}
\label{eq:freq-agg}
\mathbf{z}_t = \sum_{f=1}^{F} \alpha_{tf} \left( \mathbf{W}_v \mathbf{x}_{tf} + \mathbf{p}_f \right), 
\quad
\alpha_{tf} = \text{softmax}_f \left( \frac{\mathbf{q}^\top (\mathbf{W}_k \mathbf{x}_{tf} + \mathbf{p}_f)}{\sqrt{d_q}} \right),
\end{equation}
where $\mathbf{q} \in \mathbb{R}^{F' \times d_q}$ are learnable queries, $d_q$ is the query dimension, and $\mathbf{p}_f$ denotes sinusoidal frequency encodings. This module emphasizes informative subcarriers while compressing the frequency dimension by 5--11$\times$.

To capture activity patterns independent of their absolute location in the input window, we use relative position bias in self-attention:
\begin{equation}
\mathrm{Attn}_{ij} = \mathrm{softmax}\!\left(\frac{\mathbf{q}_i^\top \mathbf{k}_j}{\sqrt{d_k}} + b_{ij}\right)\mathbf{v}_j,
\quad
b_{ij} = f_\theta(j-i),
\end{equation}
where $f_\theta$ learns embeddings over relative temporal offsets. This encourages similar representations for the same activity pattern regardless of when it occurs.

The encoder consists of adaptive frequency aggregation followed by a 6-layer Transformer with hidden dimension $d=256$ and 8 attention heads. A 3-layer decoder is used for reconstruction during pre-training. We refer to this as the Base model (5M parameters), and also study Small (2M) and Large (12M) variants in Section~\ref{sec:scaling}.

\paragraph{End-to-end data flow.}
Given a preprocessed CSI segment $\mathbf{X} \in \mathbb{R}^{B \times 1 \times T \times F}$ with $T{=}500$ and $F{=}112$, AmbientFM first applies adaptive frequency aggregation with $F'{=}10$ learnable frequency queries for each of $c{=}16$ output channels, producing $\mathbb{R}^{B \times 16 \times T \times 10}$. The channel and frequency dimensions are then flattened and projected to frame-level embeddings in $\mathbb{R}^{B \times T \times 256}$. A 6-layer Transformer encoder with relative position bias produces the shared representation $\mathbf{Z} \in \mathbb{R}^{B \times T \times 256}$. All pre-training objectives and downstream tasks operate on $\mathbf{Z}$. During pre-training, $\mathcal{L}_{\text{CL}}$ is applied to temporally pooled $\mathbf{Z}$ with a 2-layer MLP projector, $\mathcal{L}_{\text{REC}}$ reconstructs the input CSI with a 3-layer Transformer decoder, and $\mathcal{L}_{\text{ACF}}$ predicts 50 autocorrelation lags from pooled $\mathbf{Z}$. For downstream tasks, $\mathbf{Z}$ is fed to a 1-layer LSTM and a task-specific head, with the encoder either frozen for temporal probing or adapted with bottleneck adapters.

\subsection{Self-Supervised Learning Framework}
\label{sec:ssl}

We pre-train AmbientFM with three complementary objectives: contrastive learning, masked reconstruction, and physics-informed ACF prediction. Together, they encourage the encoder to capture discriminative, robust, and temporally meaningful structure from unlabeled CSI.

\textbf{Contrastive learning.}
Given two augmented views $\mathbf{x}_1, \mathbf{x}_2$ of the same input $\mathbf{x}$, we obtain temporally pooled representations $\mathbf{z}_1, \mathbf{z}_2 \in \mathbb{R}^d$ and apply the NT-Xent loss:
\begin{equation}
\mathcal{L}_{\text{CL}} = -\frac{1}{2B} \sum_{i=1}^{2B} \log \frac{\exp(\text{sim}(\mathbf{p}_i, \mathbf{p}_{j(i)})/\tau)}{\sum_{k=1}^{2B} \exp(\text{sim}(\mathbf{p}_i, \mathbf{p}_k)/\tau)},
\end{equation}
where $\mathbf{p} = g(\mathbf{z})$ is an MLP projection, $j(i)$ denotes the positive pair, and $\tau = 0.3$. This objective encourages representations that preserve activity semantics while remaining robust to nuisance variation in channel conditions.

We use nine domain-informed augmentations that mimic realistic wireless perturbations, including additive noise, amplitude modulation, temporal shift, frequency perturbation, temporal crop, frequency masking, and Gaussian blur (Appendix~\ref{app:model_architecture}, Table~\ref{tab:signal_aug}). Each view is formed by independently sampling a subset of these transformations.

\textbf{Masked reconstruction.}
We randomly mask 10\% of temporal blocks and 10\% of frequency bands to simulate packet loss and frequency-selective fading. A reconstruction head then predicts the original signal:
\begin{equation}
\mathcal{L}_{\text{REC}} = \|\hat{\mathbf{X}} - \mathbf{X}\|_2^2.
\end{equation}
Unlike standard masked autoencoding, we reconstruct the full signal rather than only the masked regions. This encourages the model to capture temporal coherence and cross-frequency structure induced by multipath propagation.

\textbf{Physics-informed ACF prediction.}
To encourage sensitivity to motion-related temporal structure, we add an auxiliary objective that predicts the normalized autocorrelation function (ACF) of CSI energy over time. Let $s_t = \mathbb{E}_f[\mathbf{X}_{tf}^2]$. The ACF is defined as
\begin{equation}
\mathrm{ACF}[k] = \frac{\sum_{t=0}^{T-k-1} s_t s_{t+k}}{\sum_{t=0}^{T-1} s_t^2},
\end{equation}
which is computed efficiently via FFT. An MLP head predicts the first $K = 50$ lags:
\begin{equation}
\mathcal{L}_{\text{ACF}} = \|\widehat{\mathrm{ACF}} - \mathrm{ACF}(\mathbf{X})\|_2^2.
\end{equation}
This objective encourages the encoder to capture rhythmic and periodic temporal patterns associated with human activity.

The full pre-training objective is
\begin{equation}
\mathcal{L} = \lambda_{\text{CL}}\mathcal{L}_{\text{CL}} + \lambda_{\text{REC}}\mathcal{L}_{\text{REC}} + \lambda_{\text{ACF}}\mathcal{L}_{\text{ACF}},
\end{equation}
with $\lambda_{\text{CL}} = 1$, $\lambda_{\text{REC}} = 4$, and $\lambda_{\text{ACF}} = 1$. We assign a larger weight to $\mathcal{L}_{\text{REC}}$ because its raw magnitude is smaller than the other two losses, and this weighting keeps the optimization balanced.

\subsection{Pre-training Details}
\label{sec:training}

We optimize the model with AdamW using a learning rate of $5 \times 10^{-5}$ and weight decay of $10^{-4}$. Training uses a 10-epoch linear warmup followed by cosine annealing. We train in mixed precision (FP16) on 8 A100 GPUs with a total batch size of 80. To stabilize optimization, we apply gradient clipping with max norm 1.0 together with standard numerical stabilization. Pre-training runs for 100 epochs and takes approximately 165 hours (about 7 days).

\subsection{Downstream Adaptation}
\label{sec:adaptation}
 
A key advantage of foundation models is efficient adaptation to new tasks with minimal labeled data. We evaluate two adaptation strategies that span the efficiency--flexibility tradeoff.
 
\paragraph{Temporal Probing.}
The simplest adaptation strategy freezes all encoder parameters and trains only a lightweight classifier on the sequence of frame representations. Since wireless sensing depends on temporal dynamics, we use a lightweight temporal head rather than a purely linear probe:
\begin{equation}
\hat{y} = g_\phi\left(\text{LSTM}(\mathbf{z}_1, \mathbf{z}_2, \ldots, \mathbf{z}_T)\right),
\end{equation}
where $\mathbf{z}_t \in \mathbb{R}^d$ are the frozen encoder outputs and $g_\phi$ is a classification head. This is a common choice in sequential sensing models, where a purely linear probe may underuse temporal structure in frozen representations \cite{IMU-FM}.

\paragraph{Bottleneck Adaptation.}
For more expressive adaptation, we insert lightweight adapters into each transformer layer while freezing all pre-trained weights. Each adapter uses a down-projection, GELU, and up-projection, and is added to the frozen layer output through a residual connection. The up-projection is zero-initialized so that adapters are identity at initialization. Adapters are attached in parallel to the three linear layers in each encoder block: the attention output projection and the two feed-forward projections. With $r=192$ and $L=6$, this adds 2.37M trainable parameters while keeping all 5.57M encoder parameters frozen. Both adaptation methods share the same temporal classifier, isolating the effect of adapter tuning.
 
Unless otherwise noted, all downstream models are optimized with AdamW using a learning rate of $10^{-3}$ and weight decay of $10^{-4}$. We use cosine annealing with a 10-epoch linear warmup for both adaptation settings.

%%%%%%%%%%%%%%%%%%%%%%%%%%%%%%%%%%%%%%%%%%%%%%%%%%%%%%%%%%%%%%%%%%%%%%%%%%%%%%%
% SECTION 5: DOWNSTREAM EVALUATION
%%%%%%%%%%%%%%%%%%%%%%%%%%%%%%%%%%%%%%%%%%%%%%%%%%%%%%%%%%%%%%%%%%%%%%%%%%%%%%%
\section{Downstream Evaluation}
\label{sec:eval}

We now aim to understand whether a single AmbientFM backbone can provide transferable representations for diverse sensing tasks.

\subsection{Tasks and Evaluation Protocol}
\label{sec:eval_protocol}

\textbf{Tasks and Datasets.}
We evaluate AmbientFM on 9 tasks spanning 6 categories of ambient perception. 
\textit{Physiological sensing} includes fall detection, which identifies acute events from transient signal disruptions. 
\textit{Activity understanding} includes human activity recognition (HAR) and motion source recognition (MSR), which distinguishes human from non-human motion~\cite{WiMoID,SrcSense}. 
\textit{Spatial reasoning} covers occupancy detection, localization, and proximity estimation. 
\textit{Biometrics} includes user identification (UID), which distinguishes individuals based on their characteristic motion patterns. 
\textit{Fine-grained motion} is represented by gesture recognition, a 276-class task involving short-duration micro-motions. 
\textit{Spatial reconstruction} includes WiFi imaging, which reconstructs spatial structure from reflected signals. 
Together, these tasks vary in temporal scale, spatial resolution, and output format, providing a broad evaluation of representation quality.

We evaluate on three public benchmarks: CSI-Bench~\cite{CSI-Bench}, which contains 461 hours of in-the-wild data across 26 environments, 35 users, and 16 device types; SignFi~\cite{ma2020signfi} for 276-class gesture recognition; and WiFiCam~\cite{li2020wificam} for spatial reconstruction. Pre-training and downstream evaluation data are fully disjoint in device, environment, time, and participant pool, and no downstream data is seen during pre-training. 
We use the original train/test split for each benchmark. We release the split files and downstream evaluation code where source licenses permit. 

\textbf{Experiment Set-Up.}
We evaluate AmbientFM with two adaptation protocols, \textit{Temporal Probe (TP)} and \textit{Bottleneck Adaptation (BA)}, against a supervised baseline. 
TP trains only a temporal classifier on top of a frozen pre-trained encoder. 
BA keeps the encoder frozen and trains lightweight adapters with a task head. 
The \textit{supervised baseline} uses the same encoder architecture trained from scratch with task labels. 
All methods use identical data splits and downstream training settings.

Our evaluation targets \emph{cross-task} transfer from a single backbone.
We report AUROC with 95\% bootstrap confidence intervals for classification, and SSIM and PSNR for reconstruction. 
AUROC is the primary metric because it is threshold-independent and robust to class imbalance. 
Accuracy, F1, and false alarm rate are provided in Appendix~\ref{app:cross_task_transfer_results}, Table~\ref{tab:test_all_tasks_F1}.

\subsection{Cross-Task Transfer Performance}
\label{sec:crosstask_transfer}

Table~\ref{tab:crosstask_transfer} summarizes AmbientFM across 9 downstream tasks using a \emph{single} pre-trained backbone. With bottleneck adaptation, AmbientFM achieves AUROC above 0.90 on all 8 classification tasks, spanning physiological sensing, activity recognition, spatial reasoning, and biometrics. On WiFi imaging, it reaches 0.726 SSIM and 18.87 PSNR, indicating that the learned representation also preserves spatial structure for dense reconstruction.

The supervised baseline differs from AmbientFM only in the absence of pre-training. AmbientFM + BA improves AUROC by at least 0.36 on six of the eight classification tasks; on Fall Detection and UID, where the supervised baseline is already strong, the benefit of pre-training appears with limited labels (Section~\ref{sec:data_efficiency}). The two adaptation settings further clarify how reusable the learned representation is. Temporal probing already performs strongly on UID, Motion Source Recognition, Localization, Occupancy, and HAR, suggesting that many task-relevant features are directly accessible from the frozen pre-trained representation. Bottleneck adaptation is most helpful for tasks that require finer feature recombination, most notably Gesture Recognition, where performance increases from 0.634 to 0.999. This result is consistent with the difficulty of the task: Gesture Recognition involves 276 classes and short-duration micro-motions, which are harder to separate using only a frozen encoder and a lightweight temporal head.

Overall, these results show that AmbientFM provides a single transferable backbone for diverse wireless sensing tasks, supporting both classification and reconstruction.

\paragraph{Relationship to task-specific methods.}
Task-specific models can achieve higher performance when each task is optimized independently, although AmbientFM + BA exceeds the published references on HAR and Proximity (Appendix~\ref{app:task_specific}). AmbientFM does not aim to exceed these per-task upper bounds. Instead, it offers a single pre-trained backbone that transfers across all nine tasks with lightweight adaptation, replacing nine separate training pipelines with one. This tradeoff is standard in foundation models such as wav2vec~2.0~\cite{wav2vec} and LIMU-BERT~\cite{IMU-FM}, which favor transferability and label efficiency over peak task-specific accuracy. The practical advantage is reduced task-specific design and labeled data requirements, both of which remain major bottlenecks for scalable WiFi sensing.

\begin{table*}[t]
\caption{Cross-task performance across downstream WiFi sensing tasks. AUROC and 95\% confidence intervals (CI) are reported for classification tasks and SSIM\,/\,PSNR are reported for the reconstruction task Imaging$^\dagger$. }
\label{tab:crosstask_transfer}
\centering
\small
\setlength{\tabcolsep}{4pt}
\resizebox{\textwidth}{!}{%
\begin{tabular}{@{}lc@{\hspace{6pt}}c@{\hspace{6pt}}c@{\hspace{6pt}}c@{\hspace{6pt}}c@{\hspace{6pt}}c@{\hspace{6pt}}c@{\hspace{6pt}}c@{\hspace{6pt}}c@{}}
\toprule
\textbf{Method}
& \textbf{Fall}
& \textbf{HAR}
& \textbf{Gesture}
& \textbf{UID}
& \textbf{Localization}
& \textbf{MSR}
& \textbf{Occupancy}
& \textbf{Proximity}
& \textbf{Imaging}$^\dagger$ \\
\midrule
AmbientFM + BA
& \textbf{0.919}
& \textbf{0.923}
& \textbf{0.999}
& 0.993
& \textbf{0.995}
& 0.974
& \textbf{0.989}
& \textbf{0.936}
& \textbf{0.726\,/\,18.87} \\
& {\scriptsize (0.902, 0.935)}
& {\scriptsize (0.915, 0.932)}
& {\scriptsize (0.999, 0.999)}
& {\scriptsize (0.990, 0.995)}
& {\scriptsize (0.992, 0.997)}
& {\scriptsize (0.970, 0.977)}
& {\scriptsize (0.986, 0.992)}
& {\scriptsize (0.927, 0.946)}
& \\
\midrule
AmbientFM + TP
& 0.873
& 0.915
& 0.634
& \textbf{0.996}
& 0.990
& \textbf{0.992}
& 0.989
& 0.928
& 0.722\,/\,18.43 \\
& {\scriptsize (0.851, 0.893)}
& {\scriptsize (0.906, 0.922)}
& {\scriptsize (0.623, 0.645)}
& {\scriptsize (0.994, 0.997)}
& {\scriptsize (0.986, 0.994)}
& {\scriptsize (0.991, 0.993)}
& {\scriptsize (0.985, 0.992)}
& {\scriptsize (0.919, 0.938)}
& \\
\midrule
Supervised baseline
& 0.916
& 0.527
& 0.564
& 0.995
& 0.631
& 0.593
& 0.566
& 0.494
& 0.726\,/\,18.41 \\
& {\scriptsize (0.899, 0.932)}
& {\scriptsize (0.513, 0.541)}
& {\scriptsize (0.562, 0.568)}
& {\scriptsize (0.993, 0.997)}
& {\scriptsize (0.606, 0.653)}
& {\scriptsize (0.581, 0.606)}
& {\scriptsize (0.544, 0.587)}
& {\scriptsize (0.477, 0.510)}
& \\
\bottomrule
\end{tabular}
}
% \footnotetext[1]{$^\dagger$Imaging reports SSIM\,/\,PSNR.}
\end{table*}

\subsection{Data Efficiency}
\label{sec:data_efficiency}
\begin{figure}[t] \centering \includegraphics[width=1\linewidth]{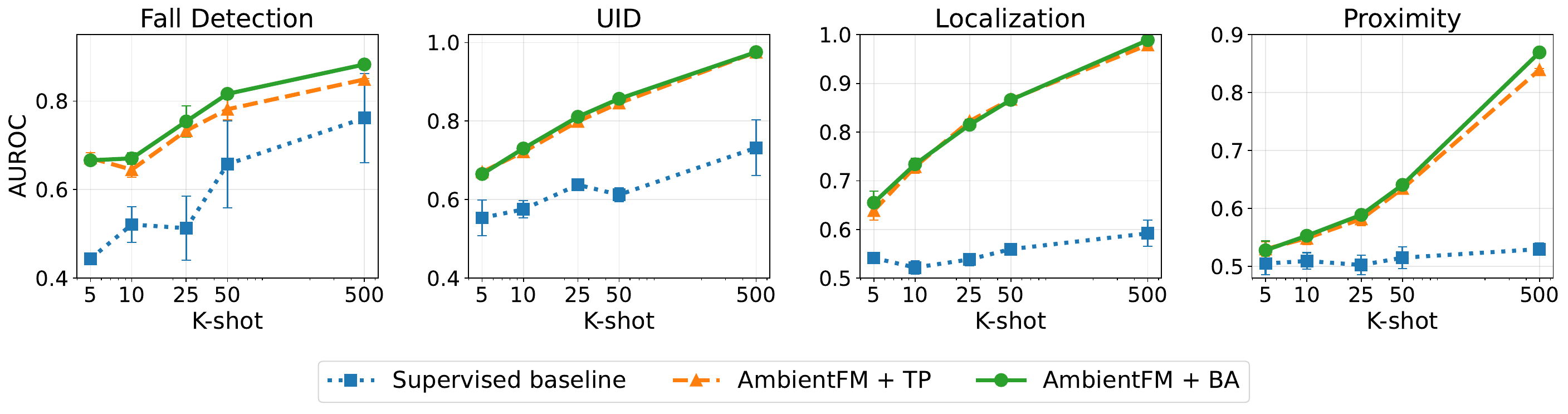} \caption{Cross-task transfer performance with small data sample sizes on four downstream tasks.} \label{fig:few-shots} \end{figure}

To assess label efficiency in realistic deployments, we restrict supervision to $K \in \{5,10,25,50,500\}$ samples per class and report mean AUROC over three seeds. Figure~\ref{fig:few-shots} shows that AmbientFM consistently outperforms training from scratch in low-data settings. With bottleneck adaptation, Fall Detection reaches 0.754 AUROC at $K=25$ and 0.816 at $K=50$, while UID reaches 0.811 at $K=25$ and 0.857 at $K=50$. The gains are especially clear for spatial tasks: Localization improves from 0.655 at $K=5$ to 0.989 at $K=500$, and Proximity improves from 0.528 to 0.870 over the same range.

These results show that AmbientFM substantially reduces the amount of task-specific labeling needed for deployment. Supervised baseline improves much more slowly, reaching only 0.592 AUROC on Localization and 0.530 on Proximity even at $K=500$, while AmbientFM reaches 0.989 and 0.870, respectively. This gap is consistent with pre-training providing transferable structure that is difficult to learn from limited labeled data alone. In practical terms, AmbientFM turns label-hungry wireless sensing tasks into tasks that can improve reliably with modest annotation effort.

\subsection{Scaling Behavior}
\label{sec:scaling}

\textbf{Model scaling.} We pre-train Small (2M), Base (5M), and Large (12M) variants under the same objective and adapt all of them with the same bottleneck protocol (Appendix~\ref{sec:app_scaling}, Table~\ref{tab:scaling}). Base delivers the best overall transfer and improves over Small on nearly every task, e.g., HAR from 0.840 to 0.923 and Proximity from 0.901 to 0.936. Large performs worse than Base on Fall, HAR, MSR, Occupancy, and Proximity while remaining strong on UID (0.990) and Localization (0.985), and this pattern persists across the tested learning-rate and warmup variations. The converged pre-training losses follow the same order: Large converges to 3.27, lower than Small (3.63) but higher than Base (3.09). Increasing capacity from Small to Base therefore improves both pre-training fit and downstream transfer, whereas further scaling under the same configuration improves neither. Base therefore provides the best trade-off under the current corpus and training configuration, and scaling larger variants with broader CSI data is a natural next step.

\textbf{Pre-training data scaling.} With the Base architecture, objectives, optimization, and adaptation protocol fixed, Fall Detection AUROC increases from 0.816 with 1/5 of the corpus to 0.884 with 3/5 and 0.919 with the full corpus, and HAR increases from 0.687 (1/5) to 0.923 (full). The 3/5 point recovers about two-thirds of the total Fall gain, consistent with diminishing returns at larger corpus sizes.

\subsection{Representation Quality}
\label{sec:repr_quality}

To understand why AmbientFM transfers well, we measure inter-class distance using the mean Euclidean distance between class centroids. Compared with randomly initialized encoders, pre-training increases class separation by 14--52$\times$ across tasks, with an average gain of 25.9$\times$ (Appendix~\ref{sec:app_repr_quality}, Table~\ref{tab:interclass}). This shows that AmbientFM learns representations where task-relevant classes are already well separated, making simple probing, bottleneck adaptation, and few-shot learning more effective.

\subsection{Discussion}
\label{sec:discussion}

\begin{figure}[t]
\centering
\includegraphics[width=1\linewidth]{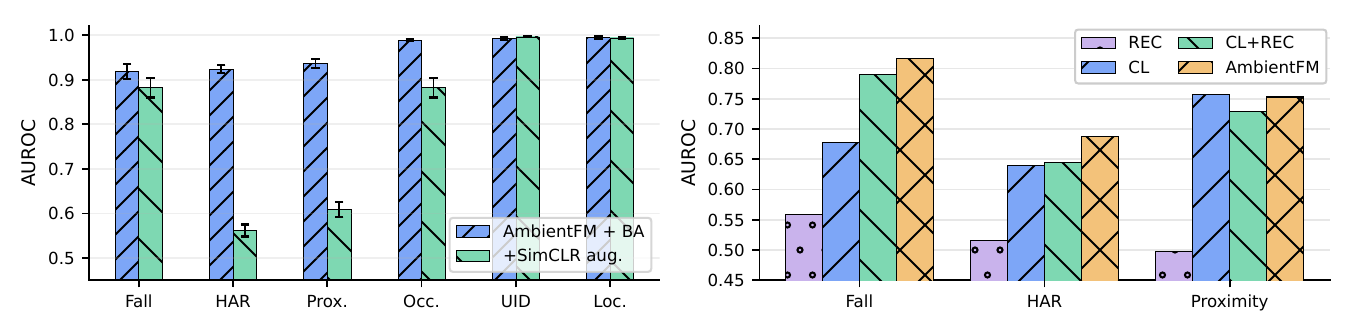}
\caption{Effect of augmentation strategy on downstream performance (left). Loss component ablation on the 1/5 pre-training corpus (right).}
\label{fig:augmentation}
\vspace{-1em}
\end{figure}

\paragraph{WiFi-tailored augmentations are critical for contrastive learning.}
Figure~\ref{fig:augmentation} (left) shows that standard SimCLR-style image augmentations are harmful when applied to CSI. Temporal tasks degrade sharply: HAR drops from 0.923 to 0.561, and Proximity from 0.936 to 0.608. In contrast, tasks dominated by stable spatial signatures are much less affected, with UID remaining at 0.993 and Localization at 0.995. This gap shows that CSI does not share the same invariances as natural images. Naive cropping or distortion can remove temporal windows and frequency components that carry sensing information. Effective contrastive pre-training therefore requires augmentations that reflect realistic wireless variation.

\paragraph{Each pre-training objective contributes to transfer quality.}
Figure~\ref{fig:augmentation} (right) ablates the pre-training objective using a 1/5 subset of the corpus, allowing us to measure the contribution of each loss component at manageable cost. The full AmbientFM objective performs best overall, especially on motion- and periodicity-related tasks: average AUROC over the three tasks is 0.524 for reconstruction-only (MAE-style) pre-training, 0.691 for contrastive-only, 0.721 for CL+REC, and 0.752 for the full objective. Removing ACF reduces Fall Detection from 0.816 to 0.790 and HAR from 0.687 to 0.645; removing reconstruction further lowers Fall Detection to 0.678. Although the 1/5-subset model has lower absolute performance than the full-corpus model, the \emph{relative ordering} across objective variants is consistent with complementary objectives: contrastive learning captures discriminative structure, reconstruction preserves signal detail, and ACF modeling improves temporal and periodic sensing.

% \begin{table}[t]
% \centering
% \small
% \caption{Loss component ablation (AUROC). Pre-training uses a 1/5 subset; downstream adaptation uses bottleneck adapters.}
% \label{tab:loss_ablation}
% \begin{tabular}{lccc}
% \toprule
% \textbf{Task} & \textbf{REC} & \textbf{CL} & \textbf{CL+REC} & \textbf{AmbientFM} \\
% \midrule
% Fall Detection & 0.559 & 0.678 & 0.790 & \textbf{0.816} \\
% HAR            & 0.515 & 0.639 & 0.645 & \textbf{0.687} \\
% Proximity      & 0.497 & \textbf{0.757} & 0.729 & 0.753 \\
% \bottomrule
% \end{tabular}
% \end{table}

\paragraph{Deployment feasibility.}
AmbientFM's Base encoder (5.57M parameters, 21.3\,MB; 4.72 GFLOPs per 500-step window) runs at 26.3--29.0\,ms per window on a commodity CPU (AMD Ryzen 7 PRO 7840U) without GPU acceleration for batch sizes 1, 8, and 32, excluding signal accumulation and the downstream LSTM and head. This is well below the 5\,s span of one input window at 100~Hz. We report the bottleneck dimension sensitivity study in Appendix~\ref{sec:bn_sensitivity} and the sensitivity to mask ratio and loss weights in Appendix~\ref{sec:app_sensitivity}.

\paragraph{Robustness and out-of-distribution transfer.}
AUROC remains above 0.92 on both 2.4 and 5~GHz for HAR and Proximity without band-specific tuning, and Fall Detection AUROC decreases gradually from 0.917 at the 100~Hz training rate to 0.813 under 40\% packet loss (Appendix~\ref{sec:app_robustness}). Under CSI-Bench's cross-user, cross-device, and cross-environment protocols, 20-shot adaptation reaches 0.704--0.808 AUROC for HAR and 0.631--0.820 for Proximity (Appendix~\ref{sec:app_ood}).
%%%%%%%%%%%%%%%%%%%%%%%%%%%%%%%%%%%%%%%%%%%%%%%%%%%%%%%%%%%%%%%%%%%%%%%%%%%%%%%
% CONCLUSION
%%%%%%%%%%%%%%%%%%%%%%%%%%%%%%%%%%%%%%%%%%%%%%%%%%%%%%%%%%%%%%%%%%%%%%%%%%%%%%%

\section{Limitations and Future Work}
\label{sec:limitations}

AmbientFM is a reusable backbone for adaptation-based WiFi sensing: each task still needs a lightweight adaptation stage rather than a zero-shot interface, and each prediction sees a 500-step context, so longer dependencies require a segment-level aggregator. Under the current corpus and configuration, Base is the best-performing scale, and larger variants have not yet improved transfer. The pre-training corpus covers geographically concentrated residences rather than hospitals or public spaces, and its consent terms prevent public release, which limits end-to-end reproduction of pre-training.

AmbientFM is also bounded by the physical limits of commodity WiFi: severely degraded links remove the temporal and spectral structure needed for sensing, and details beyond WiFi's resolution, such as fine visual appearance, are out of reach. Deployment factors vary jointly across the downstream environments, and zero-padding may expose device- or bandwidth-specific cues that our evaluation does not isolate. Passive inference of presence and activity also carries privacy and surveillance risks (Section~\ref{sec:broader_impact}). Future work includes prompt-based or zero-shot task specification~\cite{CSI_TextBench}, controlled one-factor and session-level evaluation, calibrated phase modeling, architectural sensitivity studies, and broader corpora and reconstruction benchmarks.

\section{Conclusion}
\label{sec:conclusion}

We present AmbientFM, a foundation model for ambient intelligence via WiFi sensing. Pre-trained on the largest real-world WiFi CSI corpus reported to date, 439 days of unlabeled CSI data spanning 20 device types and 11 environments, AmbientFM transfers effectively across nine diverse tasks using a single unified backbone, exceeding 0.90 AUROC on all classification tasks. Few-shot experiments demonstrate practical data efficiency. Our results establish that the foundation model paradigm is applicable to ambient intelligence: domain-specific architectures and physics-aware objectives enable learning general-purpose representations from unlabeled data, replacing task-specific pipelines with a unified approach to privacy-preserving, always-on perception.

\section{Broader Impact}
\label{sec:broader_impact}

AmbientFM points toward practical ambient sensing systems that can operate on WiFi infrastructure already present in many indoor environments. Unlike cameras or microphones, WiFi CSI does not record identifiable visual or audio content, which gives it a privacy advantage for always-on sensing. This makes WiFi-based ambient intelligence especially relevant for assistive applications such as elderly care, fall detection, respiration monitoring, and smart-home automation.

At the same time, WiFi sensing can still infer human presence, motion, and activity, so responsible deployment is essential. Systems based on AmbientFM should be used with informed consent, clear communication about what is being sensed, and compliance with applicable privacy regulations. Future deployments should also consider safeguards that limit misuse, such as purpose-specific data collection, access control, and privacy-preserving data handling. For the same reason, pre-trained checkpoints are available only upon request and approval.

\newpage
\bibliography{reference}
\bibliographystyle{plain}

%%%%%%%%%%%%%%%%%%%%%%%%%%%%%%%%%%%%%%%%%%%%%%%%%%%%%%%%%%%%

\newpage
\appendix

\section{Dataset Description}
\label{app:dataset}
 
\subsection{Pre-Training Dataset}
The pre-training dataset was designed for large-scale self-supervised representation learning, without requiring any manual annotations. CSI was collected continuously, 24/7, across diverse residential environments, capturing naturally occurring WiFi dynamics arising from everyday human activities, periods of inactivity, and environmental variations. IoT devices are distributed throughout participants' homes and operate under normal usage conditions, allowing the dataset to reflect realistic, in-the-wild sensing scenarios rather than segmented or scripted activities.
 
The resulting dataset spans 26 users, 20 IoT device types (33 devices in total), and 11 distinct residential environments, capturing heterogeneous WiFi sensing conditions across both commodity IoT deployments and dedicated sensing devices. In total, the corpus comprises over 439 days of continuous CSI recordings, amounting to approximately 20 TB of raw CSI data, $9{,}206{,}316$ data samples, collected across the 2.4~GHz and 5~GHz bands with channel bandwidths of 20/40/80~MHz.

Table~\ref{tab:corpus_compare} compares the corpus with representative public WiFi sensing datasets. It additionally spans 33 physical devices, $1{\times}1$ to $2{\times}2$ MIMO, more than 20~dB received-power variation, and 6--9 concurrent links per residence.

\begin{table}[ht]
\centering
\small
\caption{Comparison of the AmbientFM pre-training corpus with representative public WiFi sensing datasets.}
\label{tab:corpus_compare}
\resizebox{\linewidth}{!}{%
\begin{tabular}{@{}lllccccl@{}}
\toprule
\textbf{Dataset} & \textbf{Collection} & \textbf{Reported size} & \textbf{Duration} & \textbf{Devices/chipsets} & \textbf{Env.} & \textbf{Subjects} & \textbf{Band/BW} \\
\midrule
SignFi~\cite{ma2020signfi} & Scripted & 8,280 & -- & 1/1 & 2 & 5 & 5~GHz/20~MHz \\
WiFiCam~\cite{li2020wificam} & Scripted & 18,148 & 10 min & 1/1 & 1 & 1 & 2.4~GHz/20~MHz \\
Widar3.0~\cite{Widar3.0} & Scripted & 17K & -- & 1/1 & 3 & 16 & 5.8~GHz/20~MHz \\
XRF55~\cite{XRF55} & Scripted & 42.9K & 59.6 h & 1/1 & 4 & 39 & 5~GHz/20~MHz \\
CrossSense~\cite{CrossSense} & Scripted & 606K & -- & 1/1 & 3 & 100 & 5~GHz/20~MHz \\
CSI-Bench~\cite{CSI-Bench} & In-the-wild & 231.6K & 461 h & 16 configurations & 26 & 35 & 2.4/5~GHz, 20--80~MHz \\
\midrule
AmbientFM corpus & 24/7 unscripted, unlabeled & 9.2M & 439 days & 20 types / 8 families & 11 & 26 & 2.4/5~GHz, 20/40/80~MHz \\
\bottomrule
\end{tabular}}
\end{table}

\subsection{Downstream Datasets}
\label{app:downstream_datasets}
 
We evaluate AmbientFM on nine downstream tasks drawn from three public benchmarks: CSI-Bench~\cite{CSI-Bench} (CC BY-NC-ND 4.0), SignFi~\cite{ma2020signfi} (custom non-commercial research license from the College of William \& Mary), and WiFiCam~\cite{li2020wificam} (CC BY 4.0). Pre-training and downstream evaluation data are fully disjoint in device, environment, time, and participant pool. Table~\ref{tab:downstream_summary} summarizes each task; full collection protocols, hardware specifications, and data statistics are documented in the respective benchmark papers.
 
\begin{table}[ht]
\centering
\small
\caption{Summary of downstream evaluation tasks.}
\label{tab:downstream_summary}
\begin{tabular}{llccl}
\toprule
\textbf{Task} & \textbf{Source} & \textbf{\#Classes} & \textbf{\#Samples} & \textbf{Category} \\
\midrule
Fall Detection       & CSI-Bench~\cite{CSI-Bench} & 2   & 6,700   & Physiological \\
Human Activity Recog.& CSI-Bench~\cite{CSI-Bench} & 5   & 41,503  & Activity \\
Motion Source Recog.  & CSI-Bench~\cite{CSI-Bench} & 4   & 60,900  & Activity \\
Occupancy Detection  & CSI-Bench~\cite{CSI-Bench} & 2   & $\sim$6,000 & Spatial \\
Localization         & CSI-Bench~\cite{CSI-Bench} & 6   & 7,062   & Spatial \\
Proximity Recognition& CSI-Bench~\cite{CSI-Bench} & 4   & 64,154  & Spatial \\
User Identification  & CSI-Bench~\cite{CSI-Bench} & 6   & 20,300  & Biometrics \\
Gesture Recognition  & SignFi~\cite{ma2020signfi}  & 276 & 8,280   & Fine-grained motion \\
WiFi Imaging         & WiFiCam~\cite{li2020wificam}& --  & 18,148  & Reconstruction \\
\bottomrule
\end{tabular}
\end{table}
 
SignFi~\cite{ma2020signfi} captures wireless signal variations induced by 276 American Sign Language gestures from five users across a laboratory and a home setting, yielding 8,280 samples of 200 time steps $\times$ 30 subcarriers $\times$ 3 antenna pairs. WiFiCam~\cite{li2020wificam} provides temporally synchronized CSI and RGB image pairs collected via an ESP32-S3-based transmitter--receiver system at 2.4~GHz (100~Hz CSI, 30~Hz video at $640{\times}480$), containing ten minutes of continuous human walking data (18,148 aligned CSI--image pairs).

%===================================================================
\section{Data Collection Protocol}
\label{app:data_collection}
 
\subsection{Subjects and Scenarios}
\label{app:subjects}
Our pre-training dataset includes CSI data collected from 26 participants aged 5--65 years across multiple residential environments, each with distinct household compositions. The participants comprise 8 adult males, 13 adult females, and 5 children, with heights ranging from 130 to 180~cm and body weights from 30 to 90~kg. All participants signed a consent form prior to participation. The data is collected with IRB approval. Downstream data are from publicly available datasets collected under their respective IRB approvals.

\subsection{Devices and Hardware Diversity}
\label{app:device_summary}
 
To ensure comprehensive coverage of real-world IoT infrastructure, we deploy 20 commercial WiFi-enabled edge devices operating on both 2.4~GHz and 5~GHz bands with 20, 40, and 80~MHz bandwidths. The devices span eight chipset families from eight vendors---MediaTek (MT7697N, MT8516, MT8183, MT6625L, MT6625LN, MT7658CSN), Qualcomm (IPQ4019 and others), Broadcom (BCM4345), Espressif (ESP8266/ESP8285, ESP32-S3), NXP (88W8997), Marvell (88W8887), RealTek (RTL8720DN), and Altobeam (ATBM6062)---covering configurations from low-cost smart plugs to high-performance routers and smart speakers. Antenna setups range from $1{\times}1$ to $2{\times}2$ MIMO, and the device set includes Amazon Echo variants, Google Nest/Home products, Apple HomePod, Wyze cameras and plugs, and several other IoT devices. Before data collection, each device was validated with an in-house CSI verification tool for signal consistency, sampling stability, and amplitude dynamics; only devices meeting quality thresholds were deployed~\cite{WhatYouNeed,zhu2025experience}. All devices connect through an Asus GT-AX11000 router (Broadcom 4912 quad-core CPU at 2~GHz, 1~GB RAM).
 
% NOTE: Table removed to avoid redundancy with downstream benchmark documentation.
% The device list above is the pre-training corpus; downstream benchmarks use
% their own device selections as documented in their respective papers.

\subsection{Environments}
\label{app:env}
 
Data is collected across 11 distinct real-world residential environments spanning studio apartments (as small as 576~sqft), one- and two-bedroom apartments, townhouses, and multi-floor single-family houses (up to 3,800~sqft). The environments range from single-occupant studios to multi-generational households with up to six residents (including children). Floor counts range from one to three, and layouts include open-plan, partitioned, and multi-room configurations. These environments vary in wall materials, furniture density, and ambient interference, introducing rich multipath and occlusion effects. Each environment features a distinct device deployment configuration determined by user availability and spatial constraints, with the number of concurrent sensing links ranging from 6 to 9.
 
% NOTE: Table removed; the pre-training environments are distinct from
% those in any downstream benchmark.

\subsection{Continuous Data Recording}
\label{app:continuous_recording}
 
We design a scalable data acquisition framework for continuous, 24/7 collection of WiFi CSI in real residential environments. The framework operates transparently within existing home networks, capturing CSI during normal WiFi usage without modifying end devices or disrupting user behavior.
 
CSI measurements are obtained through collaboration with router chipset vendors, who provide firmware-level access and vendor-specific utilities for CSI extraction at the driver level. We implement custom acquisition software for both Linux- and FreeRTOS-based platforms that directly interfaces with WLAN kernel modules to stream CSI from all associated devices. Data is periodically uploaded to cloud storage via AWS S3 APIs with adaptive scheduling to balance network load. All CSI files are indexed using UTC-based timestamps embedded at creation time, ensuring consistent temporal referencing across deployments. This always-on pipeline captures a wide range of real-world WiFi dynamics; the resulting CSI data are naturally unlabeled and used directly for self-supervised pre-training.
 
To support optional post-hoc analysis, participants may voluntarily log coarse-grained activities (e.g., presence changes, sleep cycles, room usage) via a lightweight spreadsheet-based annotation interface. These annotations are not required for pre-training but enable investigation of whether observed signal variations correspond to user activities, environmental changes, or network instability.
 
% NOTE: Annotation tool screenshot removed; the tool is used only for
% optional post-hoc analysis and is not central to the pre-training pipeline.

%===================================================================
\section{CSI Quality Verification and Filtering Pipeline}
\label{app:csi_verification}
 
Raw CSI measurements from commodity WiFi devices exhibit substantial variability arising from differences in chipset design, firmware behavior, antenna configuration, and deployment conditions. Typical failure modes include unstable sampling intervals, timestamp drift, gain fluctuations, and limited motion sensitivity. When such signals are indiscriminately included, these artifacts can bias self-supervised objectives and encourage shortcut learning, ultimately degrading representation quality and cross-domain generalization.
 
To construct a reliable pre-training dataset, we perform structured screening at the level of each \emph{sensing device} prior to large-scale data ingestion. Rather than filtering individual segments during training, we evaluate whether each sensing device is suitable for inclusion based on the following criteria.
 
\subsection{Verification Criteria}
Inspired by existing research in CSI qualification~\cite{WhatYouNeed}, we evaluate each sensing device along four complementary dimensions. \emph{Timestamp consistency}: we analyze inter-packet intervals to estimate the effective sampling rate and its temporal stability, excluding devices with sustained deviation from the nominal rate (30 or 100~Hz), pronounced jitter, or long temporal gaps. \emph{Empty-scene stability}: for segments recorded without human motion, we assess whether CSI amplitude remains temporally stationary, rejecting devices that exhibit abnormal fluctuations, drift, or bursty variations. \emph{Motion leakage under empty conditions}: we further screen empty segments for unintended motion signatures using lightweight amplitude-derived statistics, removing devices that repeatedly exhibit motion-like patterns in supposedly static scenes. \emph{Motion sensitivity in active scenes}: for segments with known human motion, we verify that the device produces reliable motion-induced amplitude deviations relative to the corresponding empty baseline, excluding devices with insufficient sensitivity.
 
Only sensing devices satisfying all four criteria are retained for pre-training data collection.
 
\subsection{Implementation}
 
The screening procedure is implemented as an offline device-selection stage prior to large-scale data collection, supported by a custom MATLAB-based diagnostic tool. For each candidate device, short calibration sequences are recorded under both empty and motion conditions, from which the tool extracts diagnostic statistics covering temporal stability, amplitude stationarity, and motion sensitivity. The tool provides both quantitative summaries and visual diagnostics---including amplitude time series, time--subcarrier heatmaps, and stability score traces---enabling efficient identification of unstable devices, interference-dominated channels, and devices with insufficient motion sensitivity. Device-specific statistics are compared against empirically determined thresholds derived from deployment-wide distributions. The screening process is applied once per sensing device, requires no manual annotation or task-specific supervision, and ensures that retained devices exhibit both stable empty behavior and sufficient sensitivity to human motion.
 
% NOTE: MATLAB GUI and device-quality visualization figures removed to
% reduce redundancy; the diagnostic outputs are described above.

\subsection{CSI Pre-processing}
\label{app:preprocessing}
 
CSI captures the frequency-domain characteristics of wireless channel propagation. We extract CSI amplitude from raw complex-valued measurements and apply a standardized pre-processing pipeline to ensure consistency between pre-training and downstream task evaluation.
 
\subsubsection{Amplitude Extraction}
 
The foundation model operates on CSI amplitude extracted from IEEE 802.11n/ac WiFi packets. Given the complex-valued channel frequency response $H(f_k) = |H(f_k)| e^{j\phi(f_k)}$, we retain only the amplitude $A_k = |H(f_k)|$. Amplitude is preferred over complex-valued input because phase measurements are highly sensitive to carrier frequency offset, sampling clock offset, and packet detection timing---all of which vary substantially across WiFi hardware---whereas amplitude provides more stable and hardware-invariant representations.
 
\subsubsection{Data Format Standardization}
 
CSI data from different sources arrive in varying tensor layouts, including single-sample tensors of shape $(F, T, 1)$, batch tensors of shape $(N, T, F)$, and direct time--frequency matrices of shape $(T, F)$. Our pipeline automatically detects the input format based on tensor dimensions and converts all data to the canonical format $\mathbf{X} \in \mathbb{R}^{1 \times T_{\text{target}} \times F_{\text{target}}}$, where $T_{\text{target}} = 500$ time steps and $F_{\text{target}} = 112$ frequency bins corresponding to the maximum number of OFDM subcarriers across supported WiFi configurations.
 
% NOTE: Input format table removed; the three supported layouts are
% described in the paragraph above.
 
\subsubsection{Pre-processing Pipeline}
 
The pre-processing pipeline applies a strict sequence of operations to ensure consistency between pre-training and downstream evaluation. The order of operations is critical: padding must occur before normalization to match the statistics computed during pre-training.
 
\paragraph{Step 1: Shape Transformation.}
The input tensor is converted to $(T, F)$ format using format-specific transformations. Two-dimensional inputs pass through unchanged; three-dimensional inputs with trailing singleton dimensions undergo squeeze and transpose operations; batch tensors are indexed to extract individual samples.
 
\paragraph{Step 2: Zero Padding.}
Dimensions are standardized to the target shape using zero padding:
\begin{equation}
\mathbf{X}_{\text{padded}}[t, f] = 
\begin{cases}
\mathbf{X}[t, f) & \text{if } t < T_{\text{input}} \text{ and } f < F_{\text{input}} \\
0 & \text{otherwise}
\end{cases}
\end{equation}
The temporal dimension is padded or truncated to $T_{\text{target}} = 500$, and the frequency dimension is padded or truncated to $F_{\text{target}} = 112$.
 
\paragraph{Step 3: Normalization.}
Sample-wise min-max normalization is applied to the padded tensor:
\begin{equation}
\mathbf{X}_{\text{norm}} = \frac{\mathbf{X}_{\text{padded}} - \min(\mathbf{X}_{\text{padded}})}{\max(\mathbf{X}_{\text{padded}}) - \min(\mathbf{X}_{\text{padded}}) + \epsilon}
\end{equation}
where $\epsilon = 10^{-8}$ prevents division by zero. Min-max normalization is preferred because the bounded output range prevents extreme activations, preserves relative amplitude variations that encode signal propagation and activity information, and accommodates hardware-specific gain control settings. The Pad $\rightarrow$ Normalize ordering is essential: reversing it would produce different normalization statistics because min-max normalization computed on the original data excludes padded zeros, and adding zeros after normalization would alter the effective value distribution.
 
\paragraph{Step 4: Channel Dimension.}
A leading channel dimension is added to produce the final output:
\begin{equation}
\mathbf{X}_{\text{final}} = \text{unsqueeze}(\mathbf{X}_{\text{norm}}, 0) \in \mathbb{R}^{1 \times 500 \times 112}
\end{equation}

\section{Training Details}
\label{app:model_architecture}

\subsection{Foundation Model Backbone Pre-training}
\label{app:pretrain_details}

The input CSI spectrogram $\mathbf{X} \in \mathbb{R}^{1 \times T \times F}$ (where $T$ is the time dimension and $F$ is the frequency/subcarrier dimension) is first processed by an adaptive downsampling encoder that learns to compress the frequency dimension via cross-attention. The encoder operates with a single input channel, using 8 intermediate channels and 16 output channels. We employ $F_{\text{out}} = 10$ learnable query vectors, each of dimension $c_{\text{mid}} = 8$, to attend over the input frequency bins. The encoder computes attention scores between learnable queries and input frequency bins.
The flattened features are projected and processed through a transformer encoder with learned relative position bias. Table~\ref{tab:transformer_config} summarizes the architectural configurations across model scales.

\begin{table}[ht]
\centering
\caption{Transformer Encoder Hyperparameters across model scales.}
\label{tab:transformer_config}
\begin{tabular}{lccc}
\toprule
\textbf{Hyperparameter} & \textbf{Small (2M)} & \textbf{Base (5M)} & \textbf{Large (12M)} \\
\midrule
Embedding dimension ($d_{\text{model}}$) & 192 & 256 & 768 \\
Encoder layers & 4 & 6 & 12 \\
Attention heads & 6 & 8 & 12 \\
Feed-forward dimension & 384 & 512 & 2048 \\
Dropout & 0.1 & 0.1 & 0.1 \\
Max sequence length & 512 & 512 & 512 \\
\bottomrule
\end{tabular}
\end{table}

\subsubsection{Pre-training Objectives}

The backbone is pre-trained with three joint objectives specifically designed for WiFi CSI data.

We employ a SimCLR-style contrastive learning framework adapted for the unique characteristics of WiFi CSI data. Unlike natural images, CSI spectrograms exhibit strong temporal correlations, frequency-domain structure from multipath propagation, and sensitivity to environmental dynamics. Our augmentation pipeline is specifically designed to preserve semantic content while creating diverse views.

Table~\ref{tab:signal_aug} details the signal-domain augmentations.

\begin{table}[ht]
\centering
\small
\caption{Contrastive augmentations applied during pre-training. The first eight are each applied independently with $p=0.5$; time reversal is applied with $p=0.2$.}
\label{tab:signal_aug}
\begin{tabular}{llll}
\toprule
\textbf{Augmentation} & \textbf{Parameters} & \textbf{$p$} & \textbf{Rationale} \\
\midrule
Additive Gaussian Noise & $\sigma = 0.005$ & 0.5 & Receiver thermal noise \\
Amplitude Modulation & factor $= 0.05$ & 0.5 & AGC and path-loss fading \\
Frequency Perturbation & scale $= 0.02$ (FFT) & 0.5 & Hardware variation \\
Temporal Crop & crop ratio $= 0.8$ & 0.5 & Partial observation \\
Frequency Masking & mask ratio $= 0.1$ & 0.5 & Frequency-selective fading \\
Gaussian Blur & $\sigma = 1.0$, kernel $= 5$ & 0.5 & Spectral smoothing \\
Temporal Masking & mask ratio $= 0.1$ & 0.5 & Packet loss \\
Temporal Shift & max steps $= \pm 3$ & 0.5 & Asynchronous sampling \\
\midrule
Time Reversal & --- & 0.2 & Temporal invariance \\
\bottomrule
\end{tabular}
\end{table}

We employ a masked channel modeling (MCM) objective that reconstructs masked portions of the CSI spectrogram. This objective is specifically designed to capture the spatial-temporal-frequency structure unique to WiFi channel measurements. The design is motivated by three considerations: (1) \textit{Multipath Structure}---WiFi CSI encodes multipath propagation where different frequency subcarriers experience correlated but distinct phase rotations; (2) \textit{Temporal Dynamics}---human activities create time-varying signatures in CSI; and (3) \textit{Cross-Domain Invariance}---masking entire rows and columns helps learn representations robust to missing or corrupted subcarriers.

\paragraph{Physics-Informed Auxiliary Task: Autocorrelation Function Prediction.}
We incorporate a physics-informed auxiliary objective based on the Autocorrelation Function (ACF) of the CSI signal power time series. This design is inspired by the theoretical foundations of acoustic and RF sensing, particularly the relationship between signal statistics and physical phenomena.

The ACF auxiliary task is motivated by several factors: (1) \textit{Doppler Signature Encoding}---the ACF captures Doppler shifts induced by human motion; (2) \textit{Motion Pattern Discrimination}---following the theoretical framework in SrcSense~\cite{SrcSense}, the ACF encodes temporal structure of activity-induced signal variations; (3) \textit{Hardware-Invariant Features}---the ACF is robust to hardware-specific gain variations; and (4) \textit{Temporal Scale Information}---the decay rate reveals characteristic time scales of activities.

\subsubsection{Pre-training Infrastructure}
\label{app:pretraining_infra}

Pre-training is conducted on 8 NVIDIA A100 40GB GPUs (AWS P4D.24xlarge instances) using PyTorch Distributed Data Parallel (DDP) with the NCCL backend. We employ mixed precision training (FP16/BF16) and gradient checkpointing to maximize memory efficiency. The total batch size is 80 samples (10 per GPU). We use the AdamW optimizer with a learning rate of $5 \times 10^{-5}$ and cosine annealing schedule over 100 epochs. Training data is staged from FSx to local NVMe storage via parallel rclone transfers to minimize I/O bottlenecks.

\subsection{Downstream Task Adaptation}
\label{app:downstream_adaptation}

\subsubsection{Temporal Probing}

Temporal probing freezes all encoder parameters and trains only the temporal classifier, testing whether pre-trained representations are separable without encoder adaptation.

\subsubsection{Bottleneck Adaptation}

For each linear layer in the transformer encoder, we add a parallel bottleneck path:
\begin{equation}
\mathbf{y} = \mathbf{W}_{\text{base}} \mathbf{x} + \mathbf{W}_{\text{up}} \cdot \sigma(\mathbf{W}_{\text{down}} \mathbf{x}),
\end{equation}
where $\mathbf{W}_{\text{base}}$ is frozen, $\mathbf{W}_{\text{down}} \in \mathbb{R}^{r \times d_{\text{in}}}$ projects to the bottleneck with $r=192$, and $\mathbf{W}_{\text{up}} \in \mathbb{R}^{d_{\text{out}} \times r}$ projects back with GELU activation. Dropout of 0.03 is applied within the adapter. The up-projection is zero-initialized so adapters act as identity at initialization.

\subsubsection{WiFi-to-image Adaptation}
For the WiFi-imaging downstream task, we adapt the architecture by replacing the original encoder introduced in WiFiCam~\cite{li2020wificam} with our pre-trained AmbientFM model and introducing a lightweight output adapter that maps the AmbientFM backbone’s embedding dimension to the target dimension required by the image generation pipeline. During training, we employ bottleneck adaptation to fine-tune the AmbientFM model.

\subsubsection{Downstream Training Configuration}

We use largely consistent training hyperparameters for both temporal probing and bottleneck adaptation to ensure a fair comparison. Both methods are optimized using AdamW with an initial learning rate of 0.001 and a weight decay of 1e-4. Training is performed for up to 100 epochs with early stopping based on validation performance (patience of 10 epochs). Gradient norms are clipped to 1.0, and a cosine annealing learning rate scheduler is applied, with the minimum learning rate set to 1\% of the initial value. Cross-entropy loss is used for all downstream classification tasks. The main difference lies in batch size: temporal probing uses a batch size of 64, while bottleneck adaptation uses smaller batches (16–32) due to higher memory consumption during partial fine-tuning.

\begin{table}[t]
\centering
\small
\setlength{\tabcolsep}{3pt}
\caption{Test-set performance across downstream tasks. AUROC is reported with 95\% CI. FAR denotes false alarm rate.}
\label{tab:test_all_tasks_F1}

\begin{tabular}{llccccc}
\toprule
\textbf{Task} & \textbf{Method} & \textbf{AUROC} & \textbf{95\% CI} & \textbf{Accuracy} & \textbf{F1} & \textbf{FAR} \\
\midrule

\multirow{3}{*}{Fall Detection}
& Supervised baseline & 0.916 & (0.899, 0.932) & 0.827 & 0.844 & 0.117 \\
& AmbientFM + TP & 0.873 & (0.851, 0.893) & 0.801 & 0.833 & 0.252 \\
& AmbientFM + BA & 0.919 & (0.902, 0.935) & 0.834 & 0.852 & 0.125 \\
\midrule

\multirow{3}{*}{Human Activity Recognition}
& Supervised baseline & 0.527 & (0.513, 0.541) & 0.520 & 0.137 & 0.200 \\
& AmbientFM + TP & 0.915 & (0.906, 0.922) & 0.765 & 0.685 & 0.072 \\
& AmbientFM + BA & 0.923 & (0.915, 0.932) & 0.800 & 0.729 & 0.059 \\
\midrule

\multirow{3}{*}{User Identification}
& Supervised baseline & 0.995 & (0.993, 0.997) & 0.943 & 0.937 & 0.014 \\
& AmbientFM + TP & 0.996 & (0.994, 0.997) & 0.933 & 0.932 & 0.018 \\
& AmbientFM + BA & 0.993 & (0.990, 0.995) & 0.936 & 0.936 & 0.017 \\
\midrule

\multirow{3}{*}{Localization}
& Supervised baseline & 0.631 & (0.606, 0.653) & 0.273 & 0.128 & 0.162 \\
& AmbientFM + TP & 0.990 & (0.986, 0.994) & 0.914 & 0.899 & 0.019 \\
& AmbientFM + BA & 0.995 & (0.992, 0.997) & 0.938 & 0.916 & 0.014 \\
\midrule

\multirow{3}{*}{Motion Source Recognition}
& Supervised baseline & 0.593 & (0.581, 0.606) & 0.522 & 0.273 & 0.217 \\
& AmbientFM + TP & 0.992 & (0.991, 0.993) & 0.936 & 0.920 & 0.027 \\
& AmbientFM + BA & 0.974 & (0.970, 0.977) & 0.886 & 0.842 & 0.049 \\
\midrule

\multirow{3}{*}{Occupancy Detection}
& Supervised baseline & 0.566 & (0.544, 0.587) & 0.506 & 0.021 & 0.002 \\
& AmbientFM + TP & 0.989 & (0.985, 0.992) & 0.955 & 0.955 & 0.048 \\
& AmbientFM + BA & 0.989 & (0.986, 0.992) & 0.958 & 0.958 & 0.043 \\
\midrule

\multirow{3}{*}{Proximity Recognition}
& Supervised baseline & 0.494 & (0.477, 0.510) & 0.284 & 0.111 & 0.250 \\
& AmbientFM + TP & 0.928 & (0.919, 0.938) & 0.749 & 0.752 & 0.085 \\
& AmbientFM + BA & 0.936 & (0.927, 0.946) & 0.796 & 0.799 & 0.069 \\

\bottomrule
\end{tabular}
\end{table}

\subsubsection{Few-Shot Learning Configuration}

For few-shot evaluation, we sample $k \in \{5, 10, 25, 50, 500\}$ examples per class from the training set. Sampling is stratified by class label to ensure balanced representation across all categories. We report results averaged over three random seeds to ensure statistical reliability. Validation is restricted to the same $k$ labeled samples per class, and hyperparameters are inherited from the full-data configuration rather than re-tuned for each $k$.

\section{Evaluation and Ablation Details}

\subsection{Cross-Task Transfer Results}
\label{app:cross_task_transfer_results}

Table~\ref{tab:test_all_tasks_F1} provides cross-task transfer performance on the test split for a single backbone. Overall, pretraining yields strong transfer across heterogeneous sensing objectives, with \emph{Bottleneck Adaptation} achieving consistently high AUROC and low false alarms on most tasks.

\paragraph{High transfer on spatial reasoning and presence sensing.}
Localization transfers exceptionally well: Bottleneck Adaptation reaches 0.995 AUROC with 0.938 accuracy, 0.916 F1, and 0.014 FAR. Occupancy Detection is similarly strong (0.989 AUROC), with 0.958 accuracy / 0.958 F1 and 0.043 FAR. Proximity Recognition also transfers well, reaching 0.936 AUROC with improved thresholded performance (0.796 accuracy, 0.799 F1) and 0.069 FAR.

\paragraph{Strong activity understanding with modest adaptation.}
For HAR, Bottleneck Adaptation improves both ranking and decision metrics, reaching 0.923 AUROC and 0.729 F1 with 0.059 FAR. Motion Source Recognition achieves the highest AUROC with Temporal Probe (0.992) and strong decision metrics (0.936 accuracy, 0.920 F1, 0.027 FAR), while Bottleneck Adaptation maintains high performance (0.974 AUROC) with a slight FAR increase (0.049).

\paragraph{Biometrics and event sensing.}
User identification is near-saturated across all adaptation strategies, with AUROC $\geq$ 0.993 and very low FAR ($\leq$ 0.018). For Fall Detection, Bottleneck Adaptation slightly improves AUROC over supervised baseline (0.919 vs.\ 0.916) with comparable thresholded performance (0.852 F1) and low FAR (0.125).

\paragraph{Supervised baseline vs.\ pre-trained.}
Supervised baseline underperforms on most tasks, with particularly large gaps on HAR (0.527 AUROC, 0.137 F1), Localization (0.631 AUROC, 0.128 F1), and Proximity (0.494 AUROC, 0.111 F1). In contrast, temporal probing already yields strong transfer on several tasks (e.g., Localization 0.990 AUROC; Motion Source Recognition 0.992 AUROC), while Bottleneck Adaptation provides the best or most consistent improvements in thresholded metrics (e.g., HAR and Proximity) and maintains low false alarm rates.

\subsection{Comparison with Task-Specific References}
\label{app:task_specific}

Table~\ref{tab:task_specific} compares AmbientFM + BA with published task-specific references, each of which optimizes a separate model for one task. AmbientFM + BA exceeds the references on HAR (0.800 vs.\ 0.754) and Proximity (0.796 vs.\ 0.775) while using one pre-trained backbone for all nine tasks.

\begin{table}[ht]
\centering
\small
\caption{Comparison with published task-specific references. Classification tasks report accuracy and Imaging reports SSIM\,/\,PSNR. References are from CSI-Bench~\cite{CSI-Bench} for the seven CSI-Bench tasks, SignFi~\cite{ma2020signfi} for Gesture, and WiFiCam~\cite{li2020wificam} for Imaging.}
\label{tab:task_specific}
\resizebox{\linewidth}{!}{%
\begin{tabular}{@{}lccccccccc@{}}
\toprule
\textbf{Method} & \textbf{Fall} & \textbf{HAR} & \textbf{Gesture} & \textbf{UID} & \textbf{Loc.} & \textbf{MSR} & \textbf{Occ.} & \textbf{Prox.} & \textbf{Imaging} \\
\midrule
Task-specific reference & 0.949 & 0.754 & 0.948 & 0.995 & 1.000 & 0.996 & 0.988 & 0.775 & 0.749\,/\,20.39 \\
AmbientFM + BA & 0.834 & 0.800 & 0.939 & 0.936 & 0.938 & 0.886 & 0.958 & 0.796 & 0.726\,/\,18.87 \\
\bottomrule
\end{tabular}}
\end{table}

\subsection{Representation Analysis}

To evaluate the quality of learned representations, we employ t-distributed 
Stochastic Neighbor Embedding (t-SNE)~\cite{maaten2008visualizing}, a widely-used dimensionality reduction technique that projects high-dimensional feature embeddings into a two dimensional space while preserving local neighborhood structures. t-SNE visualization provides intuitive insights into how well the model separates different classes in the learned representation space. Well-separated clusters indicate that the model has learned discriminative features that capture class-specific patterns, while overlapping clusters suggest poor feature learning. We compare the quality of the representation between randomly initialized models and SSL pretrained models to demonstrate the effectiveness of pretraining.

\begin{figure}[ht]
    \centering
    % First subfigure - Random initialization
    \begin{subfigure}[b]{0.48\textwidth}
        \centering
        \includegraphics[width=\textwidth]{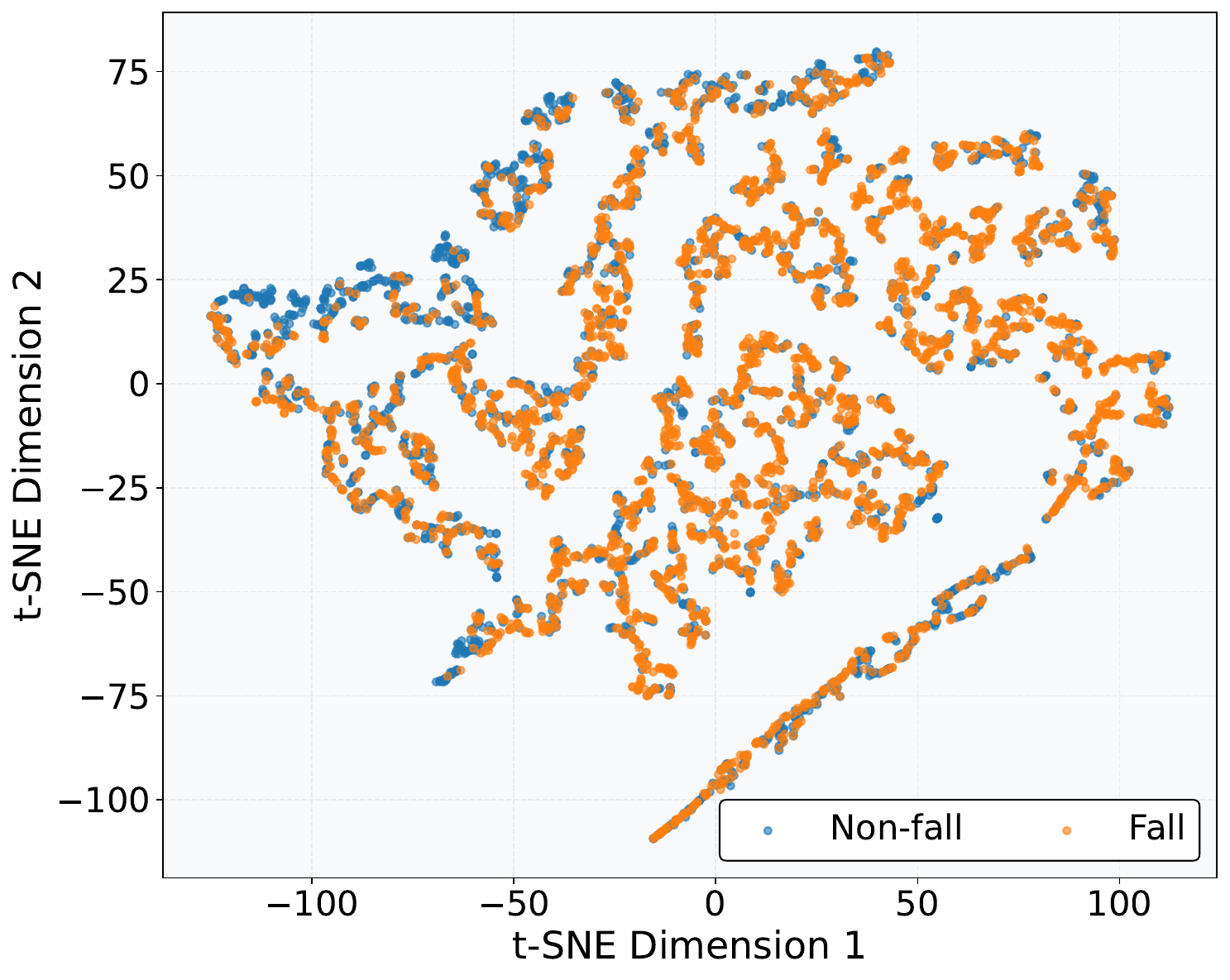}
        \caption{Random initialization}
        \label{fig:tsne_random}
    \end{subfigure}%
    \hfill
    % Second subfigure - SSL pretrained
    \begin{subfigure}[b]{0.48\textwidth}
        \centering
        \includegraphics[width=\textwidth]{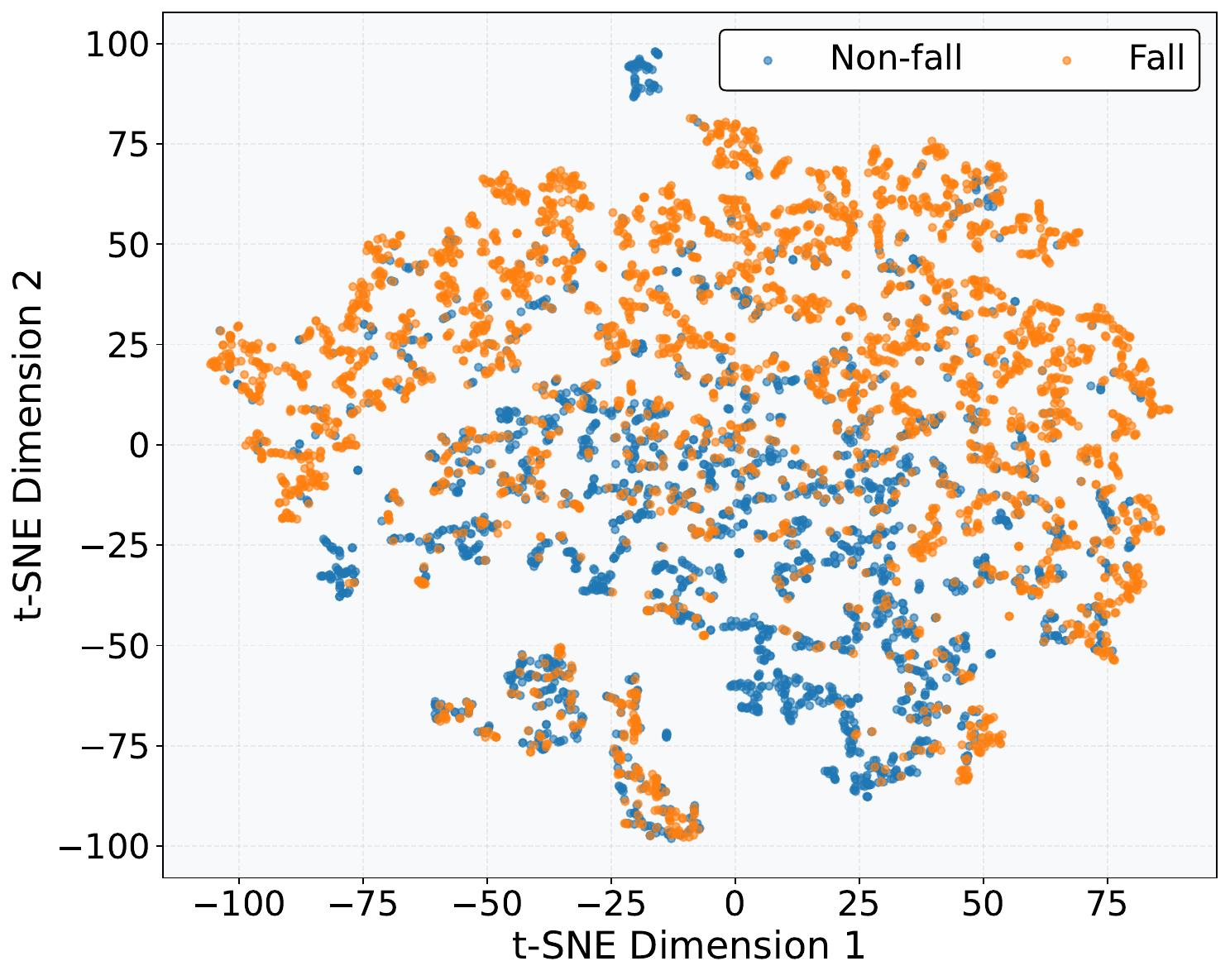}
        \caption{SSL pretrained model}
        \label{fig:tsne_pretrained}
    \end{subfigure}
    
    \caption{t-SNE visualization of learned representations for fall detection. 
    (a) Random initialization shows overlapping clusters with poor class separation, 
    indicating the model has not learned meaningful features. 
    (b) SSL pretrained model demonstrates separation between fall and non-fall classes, revealing that self-supervised pretraining learns discriminative representations beneficial for downstream tasks.}
    \label{fig:tsne_comparison}
\end{figure}

Figure~\ref{fig:tsne_comparison} illustrates the effectiveness of SSL pretraining for fall detection. The t-SNE embeddings in 
Figure~\ref{fig:tsne_random} show that randomly initialized models produce highly overlapping clusters, where fall and non-fall samples are mixed together, demonstrating the lack of meaningful feature representation. In contrast, Figure~\ref{fig:tsne_pretrained} reveals that the SSL pretrained model achieves substantially better class separation, with fall and non-fall samples forming distinct clusters in the embedding space. This clear separation indicates that the pretrained model has learned discriminative features during self-supervised pretraining that effectively capture the essential patterns that distinguish fall 
events from non-fall activities. The improved separability in the embedding space directly translates to better classification performance, validating the value of SSL pretraining for learning robust representations even without labeled data.

\begin{figure}[ht]
    \centering
    \begin{subfigure}[b]{0.48\textwidth}
        \centering
        \includegraphics[width=\textwidth]{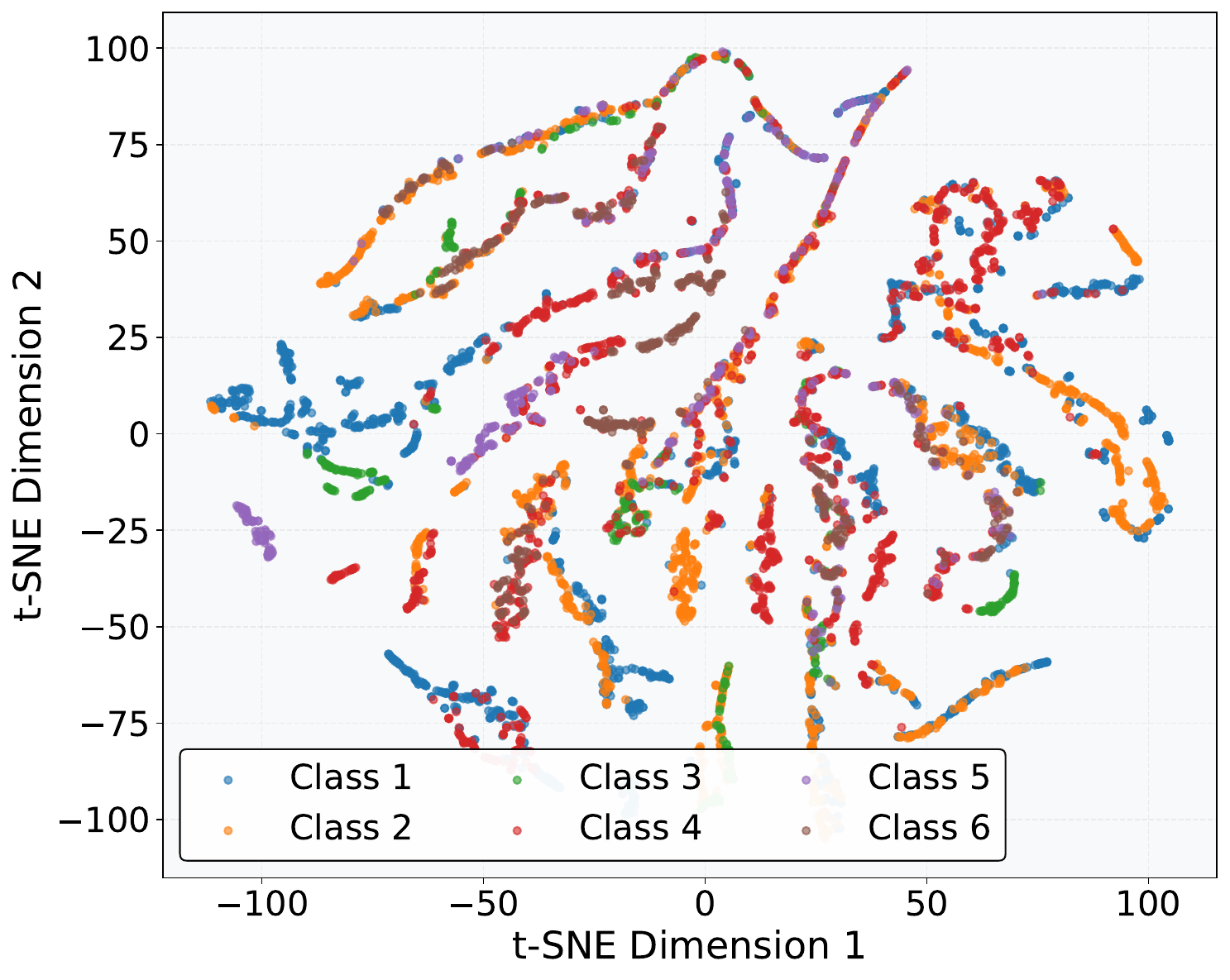}
        \caption{Random initialization}
        \label{fig:tsne_loc_random}
    \end{subfigure}%
    \hfill
    % Second subfigure - SSL pretrained
    \begin{subfigure}[b]{0.48\textwidth}
        \centering
        \includegraphics[width=\textwidth]{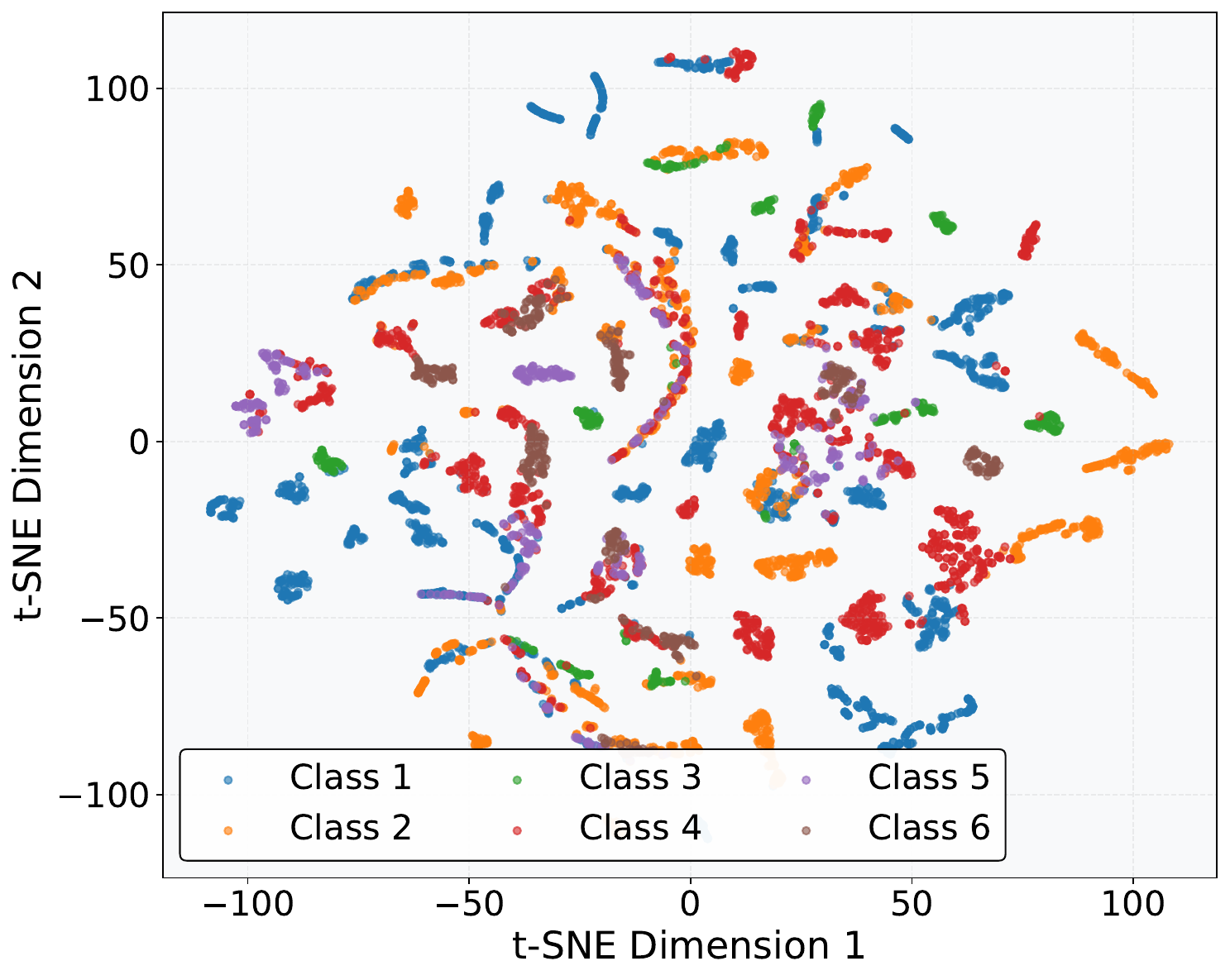}
        \caption{SSL pretrained model}
        \label{fig:tsne_loc_pretrained}
    \end{subfigure}
    
    \caption{t-SNE visualization of learned representations for localization. (a) Random initialization exhibits highly overlapping clusters across all six classes, with no discernible boundaries between different locations, indicating the absence of meaningful spatial features. (b) SSL pretrained model shows improved cluster formation with reduced inter-class overlap, demonstrating that self-supervised pretraining captures spatial patterns that facilitate location-specific classification.}
    \label{fig:tsne_localization}
\end{figure}

Figure~\ref{fig:tsne_localization} presents the t-SNE visualization comparing representation quality for the localization task. In Figure~\ref{fig:tsne_loc_random}, the randomly initialized model produces severely entangled embeddings where all data points of the six classes overlap extensively, forming intertwined patterns with no clear spatial structure. This chaotic distribution indicates that, without proper feature learning, the model cannot distinguish between different locations based on the patterns of wireless signals.

Conversely, Figure~\ref{fig:tsne_loc_pretrained} demonstrates that the SSL pretrained model achieves notably better separation in the embedding space. While some inter-class overlap persists due to the inherent complexity of the localization task, the pretrained model forms more cohesive class-specific clusters with clearer boundaries. This improvement suggests that SSL pretraining enables the model to learn location-dependent signal characteristics such as propagation patterns, multipath effects, 
and signal strength variations that are crucial for distinguishing between different locations. The enhanced separability in the pretrained embeddings translates to improved localization accuracy, validating the effectiveness of SSL pretraining for learning robust spatial representations from unlabeled wireless sensing data.

\subsection{Scaling Behavior}
\label{sec:app_scaling}

We evaluate model scaling by pre-training three variants under identical objectives and training pipelines: Small (2M parameters), Base (5M parameters), and Large (12M parameters). Architectural configurations for each variant are given in Table~\ref{tab:transformer_config}. We adapt each model using the same bottleneck protocol on CSI-Bench. Table~\ref{tab:scaling} reports the results.

The Base model achieves the best overall transfer performance. Compared to Small, it improves on most tasks: HAR increases from 0.840 to 0.923, Proximity from 0.901 to 0.936, Fall Detection from 0.898 to 0.919, Localization from 0.979 to 0.995, and Occupancy from 0.980 to 0.989. The Small model nevertheless remains competitive at less than half the parameters, with all tasks at or above 0.840 AUROC, offering an attractive trade-off for resource-constrained edge deployment.

The Large model degrades substantially on several tasks, including Fall Detection (0.627), HAR (0.524), MSR (0.515), Occupancy (0.651), and Proximity (0.496), while maintaining strong performance on UID (0.990) and Localization (0.985). This pattern persists across the tested warmup-schedule and learning-rate variations, so it is not explained by one particular hyperparameter choice. All three scales reach stable plateaus in pre-training; Large converges to a loss of 3.27, lower than Small (3.63) but higher than Base (3.09) (Table~\ref{tab:scaling}). Increasing capacity from Small to Base thus improves both pre-training fit and downstream transfer, whereas further scaling under the same configuration improves neither. Base therefore provides the best trade-off under the current corpus and training configuration. Scaling AmbientFM to larger variants is a natural direction for future work as broader CSI collection becomes available.

\begin{table}[t]
\centering
\small
\caption{Downstream performance (AUROC) across model scales, with the converged pre-training loss.}
\label{tab:scaling}
\begin{tabular}{lcccccccc}
\toprule
Model & Fall & HAR & UID & Loc. & MSR & Occ. & Prox. & Pre-train loss \\
\midrule
Small (2M) & 0.898 & 0.840 & 0.984 & 0.979 & 0.971 & 0.980 & 0.901 & 3.63 \\
Base (5M)  & 0.919 & 0.923 & 0.993 & 0.995 & 0.974 & 0.989 & 0.936 & 3.09 \\
Large (12M)& 0.627 & 0.524 & 0.990 & 0.985 & 0.515 & 0.651 & 0.496 & 3.27 \\
\bottomrule
\end{tabular}
\end{table}

\subsection{Representation Quality}
\label{sec:app_repr_quality}

\begin{table}[t]
\centering
\small
\caption{Inter-class distance in the representation space.}
\label{tab:interclass}
\begin{tabular}{lcccccc}
\toprule
Metric & Fall & Loc. & UID & Prox. & HAR & MSR \\
\midrule
Pre-trained & 3.61 & 4.12 & 5.16 & 3.64 & 1.57 & 2.58 \\
Random Init & 0.25 & 0.24 & 0.21 & 0.14 & 0.03 & 0.12 \\
Ratio ($\times$) & 14.4 & 17.4 & 24.2 & 26.5 & 52.0 & 21.0 \\
\midrule
Average & \multicolumn{6}{c}{Pre: 3.45 \quad Rand: 0.16 \quad Ratio: 25.9$\times$} \\
\bottomrule
\end{tabular}
\end{table}

To understand why AmbientFM transfers well, we quantify representation geometry using inter-class distance, defined as the mean Euclidean distance between class centroids in the encoder's output space. We compute this metric using up to 2,000 samples per task and compare pre-trained encoders against randomly initialized ones under the same architecture. Table~\ref{tab:interclass} reports the results.

Pre-training produces substantially larger class separation across all tasks, ranging from 14.4$\times$ (Fall Detection) to 52.0$\times$ (HAR), with an average gain of 25.9$\times$ despite using no downstream labels. This geometric structure has two practical consequences. First, it explains the strong temporal probing results in Section~\ref{sec:crosstask_transfer}: when classes form well-separated clusters, a simple LSTM readout on frozen representations can succeed without any encoder adaptation. Second, it explains the few-shot behavior in Section~\ref{sec:data_efficiency}: well-separated clusters require only a small number of labeled examples to place decision boundaries, making pre-training particularly valuable in low-data regimes.

The per-task separation ratios also correlate with the transfer gaps observed in Table~\ref{tab:crosstask_transfer}. HAR exhibits the highest separation gain (52.0$\times$) and the largest AUROC improvement from pre-training (0.527 to 0.923). Proximity shows 26.5$\times$ higher separation and improves from 0.494 to 0.936. Conversely, UID, which already achieves 0.995 from scratch, has a high absolute separation (5.16) but a comparatively modest ratio (24.2$\times$), consistent with the signal being inherently discriminative even without pre-training. This correspondence between representation geometry and downstream performance is consistent with pre-training organizing the encoder's feature space in a task-relevant manner.

\subsection{Bottleneck Dimension Sensitivity}
\label{sec:bn_sensitivity}

\begin{table}[t]
\centering
\small
\caption{Bottleneck dimension ($r$) trade-off on Proximity Recognition. Trainable Params includes adapter parameters across all 6 encoder layers plus the LSTM temporal classifier and task head.}
\label{tab:bn_tradeoff}
\begin{tabular}{cccc}
\toprule
\textbf{$r$} & \textbf{Trainable Params} & \textbf{Accuracy} & \textbf{AUROC} \\
\midrule
64  & 1.98M & 0.778 & 0.928 \\
128 & 2.77M & 0.787 & 0.935 \\
192 & 3.55M & 0.794 & \textbf{0.936} \\
256 & 4.34M & 0.764 & 0.932 \\
\bottomrule
\end{tabular}
\end{table}

To study the effect of adapter expressiveness, we vary the bottleneck dimension $r \in \{64, 128, 192, 256\}$ while keeping the encoder frozen, and evaluate on Proximity Recognition (Table~\ref{tab:bn_tradeoff}). Performance increases from $r=64$ to $r=192$, where accuracy (0.794) and AUROC (0.936) are highest. At $r=256$, both metrics decline, indicating that excess adapter capacity leads to overfitting with limited downstream labels. The $r=128$ configuration offers a favorable trade-off for resource-constrained deployments, achieving 0.935 AUROC with 22\% fewer trainable parameters than $r=192$.

\subsection{Sensitivity to Pre-training Hyperparameters}
\label{sec:app_sensitivity}

We vary one pre-training parameter at a time on the fixed 1/5 pre-training corpus and report HAR AUROC after bottleneck adaptation (Table~\ref{tab:sensitivity}). The default mask ratio of 0.10 improves over 0.05 and 0.25 by 0.040 and 0.012 AUROC, and the default loss weighting $\lambda_{\text{CL}}{:}\lambda_{\text{REC}}{:}\lambda_{\text{ACF}} = 1{:}4{:}1$ improves over 1:1:1 and 1:8:1 by 0.048 and 0.015. These results support the selected defaults.

\begin{table}[ht]
\centering
\small
\caption{Sensitivity of HAR AUROC to mask ratio and loss weights (1/5 pre-training corpus, bottleneck adaptation).}
\label{tab:sensitivity}
\begin{tabular}{lccc}
\toprule
\textbf{Parameter} & \textbf{Lower} & \textbf{Default} & \textbf{Higher} \\
\midrule
Mask ratio & 0.05: 0.647 & 0.10: \textbf{0.687} & 0.25: 0.675 \\
Loss weights ($\lambda_{\text{CL}}{:}\lambda_{\text{REC}}{:}\lambda_{\text{ACF}}$) & 1:1:1: 0.639 & 1:4:1: \textbf{0.687} & 1:8:1: 0.672 \\
\bottomrule
\end{tabular}
\end{table}

The remaining architectural settings balance frequency compression, token capacity, and temporal horizon. $F'{=}10$ compresses up to 112 input frequency bins into 10 latent channels; $c{=}16$ output channels give a 160-dimensional intermediate token before projection to $d{=}256$; and $K{=}50$ ACF lags cover 0.5\,s at 100~Hz.

\subsection{Robustness to Carrier Band and Packet Rate}
\label{sec:app_robustness}

Table~\ref{tab:band} reports HAR and Proximity by carrier band with the same backbone and adaptation protocol. AUROC remains above 0.92 on both bands for both tasks without band-specific tuning.

\begin{table}[ht]
\centering
\small
\caption{Performance by carrier band. Accuracy, AUROC, and F1 are mean $\pm$ standard deviation.}
\label{tab:band}
\begin{tabular}{llccccc}
\toprule
\textbf{Task} & \textbf{Band} & \textbf{Test samples} & \textbf{Accuracy} & \textbf{AUROC} & \textbf{F1} & \textbf{FAR} \\
\midrule
HAR & 2.4~GHz & 1,067 & 0.876 $\pm$ 0.025 & 0.958 $\pm$ 0.007 & 0.827 $\pm$ 0.040 & 0.039 \\
HAR & 5~GHz & 2,333 & 0.789 $\pm$ 0.013 & 0.927 $\pm$ 0.004 & 0.716 $\pm$ 0.022 & 0.068 \\
Proximity & 2.4~GHz & 527 & 0.820 $\pm$ 0.019 & 0.953 $\pm$ 0.006 & 0.821 $\pm$ 0.021 & 0.060 \\
Proximity & 5~GHz & 1,201 & 0.776 $\pm$ 0.021 & 0.933 $\pm$ 0.011 & 0.778 $\pm$ 0.022 & 0.076 \\
\bottomrule
\end{tabular}
\end{table}

To test unseen effective packet rates, we evaluate the 100~Hz Fall Detection model after randomly removing packets (Table~\ref{tab:packet_rate}). AUROC decreases gradually and remains 0.813 under 40\% packet loss.

\begin{table}[ht]
\centering
\small
\caption{Fall Detection under reduced effective packet rates. Mean $\pm$ standard deviation across repeated evaluations; the 100~Hz mean is consistent with Table~\ref{tab:crosstask_transfer} (0.919).}
\label{tab:packet_rate}
\begin{tabular}{ccccc}
\toprule
\textbf{Rate} & \textbf{Packet loss} & \textbf{Accuracy} & \textbf{AUROC} & \textbf{F1} \\
\midrule
100~Hz & 0\% & 0.839 $\pm$ 0.019 & 0.917 $\pm$ 0.009 & 0.860 $\pm$ 0.017 \\
95~Hz & 5\% & 0.802 $\pm$ 0.023 & 0.878 $\pm$ 0.032 & 0.829 $\pm$ 0.013 \\
90~Hz & 10\% & 0.789 $\pm$ 0.028 & 0.862 $\pm$ 0.045 & 0.813 $\pm$ 0.015 \\
80~Hz & 20\% & 0.760 $\pm$ 0.041 & 0.843 $\pm$ 0.058 & 0.777 $\pm$ 0.026 \\
60~Hz & 40\% & 0.720 $\pm$ 0.069 & 0.813 $\pm$ 0.100 & 0.729 $\pm$ 0.056 \\
\bottomrule
\end{tabular}
\end{table}

\subsection{Cross-User, Cross-Device, and Cross-Environment Evaluation}
\label{sec:app_ood}

The original benchmark splits do not isolate users, devices, or environments individually. We therefore additionally evaluate HAR and Proximity with CSI-Bench's explicit cross-user, cross-device, and cross-environment protocols (Table~\ref{tab:ood}). Zero-shot accuracy is above chance throughout, and at 20 shots AUROC reaches 0.704--0.808 for HAR and 0.631--0.820 for Proximity on held-out users, devices, and environments.

\begin{table}[ht]
\centering
\small
\caption{Accuracy\,/\,AUROC under CSI-Bench's cross-domain protocols. Proximity has no 10-shot setting.}
\label{tab:ood}
\begin{tabular}{llccccc}
\toprule
\textbf{Task} & \textbf{Split} & \textbf{0-shot} & \textbf{1-shot} & \textbf{5-shot} & \textbf{10-shot} & \textbf{20-shot} \\
\midrule
\multirow{3}{*}{HAR} & Cross-device & 0.411\,/\,0.604 & 0.388\,/\,0.591 & 0.483\,/\,0.676 & 0.537\,/\,0.769 & 0.575\,/\,0.808 \\
& Cross-environment & 0.425\,/\,0.537 & 0.399\,/\,0.584 & 0.400\,/\,0.615 & 0.430\,/\,0.654 & 0.492\,/\,0.704 \\
& Cross-user & 0.498\,/\,0.605 & 0.451\,/\,0.626 & 0.477\,/\,0.670 & 0.508\,/\,0.718 & 0.550\,/\,0.753 \\
\midrule
\multirow{3}{*}{Proximity} & Cross-device & 0.432\,/\,0.638 & 0.430\,/\,0.661 & 0.497\,/\,0.705 & -- & 0.605\,/\,0.820 \\
& Cross-environment & 0.357\,/\,0.600 & 0.366\,/\,0.586 & 0.405\,/\,0.618 & -- & 0.516\,/\,0.738 \\
& Cross-user & 0.303\,/\,0.558 & 0.320\,/\,0.560 & 0.323\,/\,0.571 & -- & 0.390\,/\,0.631 \\
\bottomrule
\end{tabular}
\end{table}

\paragraph{Device-level statistics.} Table~\ref{tab:group} reports device-macro metrics for HAR and Proximity, weighting each device equally. Device-macro AUROC reaches 0.920 for HAR and 0.952 for Proximity, consistent with transfer across individual devices.

\begin{table}[ht]
\centering
\small
\caption{Device-macro performance. Values are mean $\pm$ standard deviation across three seeds.}
\label{tab:group}
\begin{tabular}{lccc}
\toprule
\textbf{Task} & \textbf{Device-macro Acc.} & \textbf{Device-macro AUROC} & \textbf{Device-macro F1} \\
\midrule
HAR & 0.730 $\pm$ 0.073 & 0.920 $\pm$ 0.038 & 0.683 $\pm$ 0.088 \\
Proximity & 0.834 $\pm$ 0.076 & 0.952 $\pm$ 0.033 & 0.833 $\pm$ 0.088 \\
\bottomrule
\end{tabular}
\end{table}

%%%%%%%%%%%%%%%%%%%%%%%%%%%%%%%%%%%%%%%%%%%%%%%%%%%%%%%%%%%%

\newpage
\section*{NeurIPS Paper Checklist}

\begin{enumerate}

\item {\bf Claims}
    \item[] Question: Do the main claims made in the abstract and introduction accurately reflect the paper's contributions and scope?
    \item[] Answer: \answerYes{}
    \item[] Justification: The abstract and Section~\ref{sec:intro} state the paper's main claims, and Sections~\ref{sec:crosstask_transfer}--\ref{sec:repr_quality} provide the transfer, data-efficiency, scaling, and representation analyses that support them.

\item {\bf Limitations}
    \item[] Question: Does the paper discuss the limitations of the work performed by the authors?
    \item[] Answer: \answerYes{}
    \item[] Justification: Section~\ref{sec:limitations} discusses the need for task-specific adaptation, the 500-step context, the best-performing model scale, the representativeness and availability of the pre-training corpus, the physical sensing limits of commodity WiFi, the scope of the deployment evaluation, and privacy risks.

\item {\bf Theory assumptions and proofs}
    \item[] Question: For each theoretical result, does the paper provide the full set of assumptions and a complete (and correct) proof?
    \item[] Answer: \answerNA{}
    \item[] Justification: The paper does not present formal theoretical results such as theorems or lemmas. Equations in Sections~\ref{sec:architecture}--\ref{sec:ssl} define architectural components and training objectives rather than theoretical claims requiring proof.

    \item {\bf Experimental result reproducibility}
    \item[] Question: Does the paper fully disclose all the information needed to reproduce the main experimental results of the paper to the extent that it affects the main claims and/or conclusions of the paper (regardless of whether the code and data are provided or not)?
    \item[] Answer: \answerYes{}
    \item[] Justification: The paper specifies the architecture, objectives, preprocessing, training setup, adaptation methods, and evaluation protocol in Sections~\ref{sec:architecture}--\ref{sec:eval_protocol} and Appendices~\ref{app:preprocessing}, \ref{app:pretrain_details}, and \ref{app:downstream_adaptation}. Downstream evaluation uses public benchmarks with their original splits; the split files and downstream evaluation code are released, and pre-trained checkpoints are available upon request.

\item {\bf Open access to data and code}
    \item[] Question: Does the paper provide open access to the data and code, with sufficient instructions to faithfully reproduce the main experimental results, as described in supplemental material?
    \item[] Answer: \answerNo{}
    \item[] Justification: The pre-training corpus is not publicly released because its consent terms do not permit redistribution of in-home recordings, which include data from minors. Downstream evaluation uses publicly available benchmarks with their original splits. We release the split files and downstream evaluation code, together with environment and execution instructions, where source licenses permit. Pre-trained model weights are available to researchers upon request subject to approval, so the main transfer results can be verified on the public benchmarks.

\item {\bf Experimental setting/details}
    \item[] Question: Does the paper specify all the training and test details (e.g., data splits, hyperparameters, how they were chosen, type of optimizer) necessary to understand the results?
    \item[] Answer: \answerYes{}
    \item[] Justification: Sections~\ref{sec:training} and \ref{sec:adaptation}, Tables~\ref{tab:transformer_config} and \ref{tab:signal_aug}, and Appendix~\ref{app:downstream_adaptation} provide the key hyperparameters, optimizer, schedules, adaptation settings, and evaluation protocol.

\item {\bf Experiment statistical significance}
    \item[] Question: Does the paper report error bars suitably and correctly defined or other appropriate information about the statistical significance of the experiments?
    \item[] Answer: \answerYes{}
    \item[] Justification: Section~\ref{sec:eval_protocol} and Table~\ref{tab:crosstask_transfer} report 95\% bootstrap confidence intervals for classification tasks, and Section~\ref{sec:data_efficiency} reports few-shot results averaged over three random seeds. Appendices~\ref{sec:app_robustness} and \ref{sec:app_ood} report mean $\pm$ standard deviation across repeated evaluations, seeds, or groups, as stated in each table caption.

\item {\bf Experiments compute resources}
    \item[] Question: For each experiment, does the paper provide sufficient information on the computer resources (type of compute workers, memory, time of execution) needed to reproduce the experiments?
    \item[] Answer: \answerYes{}
    \item[] Justification: Section~\ref{sec:training} and Appendix~\ref{app:pretraining_infra} report the GPU type, training setup, batch size, and approximate total pre-training time; Section~\ref{sec:discussion} reports CPU inference cost.
    
\item {\bf Code of ethics}
    \item[] Question: Does the research conducted in the paper conform, in every respect, with the NeurIPS Code of Ethics \url{https://neurips.cc/public/EthicsGuidelines}?
   \item[] Answer: \answerYes{}
    \item[] Justification: The paper states that pre-training data collection used informed consent and IRB approval (Appendix~\ref{app:data_collection}) and discusses privacy-related risks and mitigation in Section~\ref{sec:broader_impact}.

\item {\bf Broader impacts}
    \item[] Question: Does the paper discuss both potential positive societal impacts and negative societal impacts of the work performed?
    \item[] Answer: \answerYes{}
    \item[] Justification: Section~\ref{sec:broader_impact} discusses both beneficial applications and potential harms, including privacy-sensitive inference and misuse, together with recommended safeguards.
    
\item {\bf Safeguards}
    \item[] Question: Does the paper describe safeguards that have been put in place for responsible release of data or models that have a high risk for misuse (e.g., pre-trained language models, image generators, or scraped datasets)?
    \item[] Answer: \answerYes{}
    \item[] Justification: The pre-training corpus is not released, and pre-trained checkpoints are available only upon request and approval to limit misuse of a model that can infer human presence and activity (Section~\ref{sec:broader_impact}). Data collection used informed consent and IRB approval (Appendix~\ref{app:data_collection}).

\item {\bf Licenses for existing assets}
    \item[] Question: Are the creators or original owners of assets (e.g., code, data, models), used in the paper, properly credited and are the license and terms of use explicitly mentioned and properly respected?
    \item[] Answer: \answerYes{}
    \item[] Justification: Appendix~\ref{app:downstream_datasets} cites the public downstream benchmarks and lists their licenses, and the paper states that the pre-training dataset was collected by the authors with participant consent.

\item {\bf New assets}
    \item[] Question: Are new assets introduced in the paper well documented and is the documentation provided alongside the assets?
    \item[] Answer: \answerYes{}
    \item[] Justification: The released split files and downstream evaluation code are documented with environment and execution instructions. The pre-training corpus is not released because its consent terms do not permit redistribution, and the model is available upon request; their design and collection are documented in Section~\ref{sec:pretraining_dataset} and Appendices~\ref{app:dataset}--\ref{app:csi_verification}.

\item {\bf Crowdsourcing and research with human subjects}
    \item[] Question: For crowdsourcing experiments and research with human subjects, does the paper include the full text of instructions given to participants and screenshots, if applicable, as well as details about compensation (if any)? 
    \item[] Answer: \answerYes{}
    \item[] Justification: The pre-training data collection involved 26 participants who signed informed consent forms prior to participation. Appendix~\ref{app:subjects} describes participant demographics, and Appendix~\ref{app:continuous_recording} describes the collection protocol and annotation interface. Participants voluntarily recorded coarse activity annotations via a lightweight Google Spreadsheet interface. The WiFi CSI collection is passive and non-invasive, operating transparently within existing home networks. Downstream datasets are from publicly available benchmarks collected under their respective IRB approvals.

\item {\bf Institutional review board (IRB) approvals or equivalent for research with human subjects}
    \item[] Question: Does the paper describe potential risks incurred by study participants, whether such risks were disclosed to the subjects, and whether Institutional Review Board (IRB) approvals (or an equivalent approval/review based on the requirements of your country or institution) were obtained?
    \item[] Answer: \answerYes{}
    \item[] Justification: All participants signed informed consent forms prior to data collection, and the pre-training data collection was conducted under Institutional Review Board (IRB) approval. Downstream evaluation datasets are publicly available benchmarks collected under their respective IRB approvals as discussed in Appendix~\ref{app:data_collection}.

\item {\bf Declaration of LLM usage}
    \item[] Question: Does the paper describe the usage of LLMs if it is an important, original, or non-standard component of the core methods in this research? Note that if the LLM is used only for writing, editing, or formatting purposes and does \emph{not} impact the core methodology, scientific rigor, or originality of the research, declaration is not required.
    %this research? 
    \item[] Answer: \answerNA{}
    \item[] Justification: LLMs are not part of the method, training pipeline, or evaluation. The work focuses on self-supervised learning for WiFi CSI.

\end{enumerate}

\end{document}